\documentclass{article}
\usepackage[accepted]{icml2023}

\usepackage[utf8]{inputenc} % allow utf-8 input
\usepackage[T1]{fontenc}    % use 8-bit T1 fonts
\usepackage{hyperref}       % hyperlinks
\usepackage{url}            % simple URL typesetting
\usepackage{booktabs}       % professional-quality tables
\usepackage{nicefrac}       % compact symbols for 1/2, etc.
\usepackage{microtype}      % microtypography
\usepackage{xcolor}         % colors

\usepackage{wrapfig}

\usepackage{amsfonts}       % blackboard math symbols
\usepackage{amsmath}
\usepackage{amssymb}
\usepackage{mathtools}
\usepackage{amsthm}

\usepackage{algorithm}
\usepackage{algorithmic}
\renewcommand{\algorithmiccomment}[1]{#1}

\usepackage{comment}
\usepackage{graphicx} %,subfigure}
\usepackage[capitalize,noabbrev]{cleveref}

\usepackage{bbold}

\usepackage[utf8]{inputenc}

\usepackage{verbatim}
\usepackage{ifthen}
\usepackage{xspace}
\usepackage{graphicx}
\usepackage{multimedia} % pour beamer
\usepackage{pgf}

\usepackage{amsmath,amssymb}
\usepackage{tikz}

\usepackage{color}
\usepackage{xcolor}
\usepackage{tcolorbox}
\usepackage{colortbl}
\usepackage[makeroom]{cancel}

\usepackage{ucs}
\usepackage{comment}
\usepackage{fancybox}%fait des boites
\usepackage{bm}
\usepackage{rotating}

\usepackage{array, makecell}
\usepackage{multirow}
\usepackage{colortbl}
\usepackage{tcolorbox}
\usepackage{xspace}

\newcommand{\burl}[1]{\structure{\url{#1}}}

\newcommand\no[1]{}
\newcommand\wi[1]{$\circ$}
\newcommand\bu[1]{$\bullet$}
\newcommand\ot[1]{$\star$}
\newcommand\bo[1]{$\bullet\star$}

\newcommand{\ddpg}{{\sc ddpg}\xspace}

\newcommand{\her}{\textsc{her}\xspace}

\newcommand{\ggan}{{\sc Goal gan}\xspace}

\newcommand{\rig}{{\sc rig}\xspace}

\newcommand{\diayn}{{\sc diayn}\xspace}
\newcommand{\discern}{{\sc discern}\xspace}

\newcommand{\skewfit}{{\sc Skew-Fit}\xspace}

\newcommand{\svgg}{{\sc svgg}\xspace}

\newcommand{\setso}{{\sc setter-solver}\xspace}
\newcommand{\mega}{{\sc mega}\xspace}
\newcommand{\omeg}{{\sc omega}\xspace}

\newcommand{\vds}{{\sc vds}\xspace}

\newcommand{\svgd}{{\sc svgd}\xspace} 
\newcommand{\svpg}{{\sc svpg}\xspace}

\newcommand{\argmin}{\mathop{\rm argmin}}

\newcommand{\rien}[1]{}

\DeclareGraphicsExtensions{.jpg,.mps,.pdf,.png,.gif}

\definecolor{myred}{rgb}{0.8,0,0}
\definecolor{mygreen}{rgb}{0,0.6,0}
\definecolor{myblue}{rgb}{0,0,0.7}

\definecolor{DarkGray}{gray}{0.9}
\definecolor{MediumGray}{gray}{0.75}
\definecolor{LightGray}{gray}{0.5}
\newcounter{ques} 
\newcommand{\ques}{\arabic{ques}}

\definecolor{deletecolor}{rgb}{0.5,0.5,0.5}

\newtheorem{innercustomthm}{Theorem}
\newenvironment{customthm}[1]
  {\renewcommand\theinnercustomthm{#1}\innercustomthm}
  {\endinnercustomthm}

\newtheorem{theorem}{Theorem}

\begin{document}
\icmltitlerunning{Stein Variational Goal Generation for adaptive Exploration in Multi-Goal RL}

\twocolumn[
\icmltitle{Stein Variational Goal Generation for adaptive Exploration in Multi-Goal Reinforcement Learning}

\icmlsetsymbol{equal}{*}

\begin{icmlauthorlist}
\icmlauthor{Nicolas Castanet}{isir}
\icmlauthor{Olivier Sigaud}{isir}
\icmlauthor{Sylvain Lamprier}{leria}
\end{icmlauthorlist}

\icmlaffiliation{isir}{Sorbonne Université, ISIR, Paris, France}
\icmlaffiliation{leria}{Univ Angers, LERIA, SFR MATHSTIC, F-49000 Angers, France}

\icmlcorrespondingauthor{Nicolas Castanet}{nicolas.castanet@isir.upmc.fr}
\icmlcorrespondingauthor{Olivier Sigaud}{olivier.sigaud@isir.upmc.fr}
\icmlcorrespondingauthor{Sylvain Lamprier}{sylvain.lamprier@univ-angers.fr}

\icmlkeywords{Machine Learning, ICML}
%\maketitle

\vskip 0.3in
]

\printAffiliationsAndNotice{} % otherwise use the standard text.

\begin{abstract}
In multi-goal Reinforcement Learning, an agent can share experience between related training tasks, resulting in better generalization for new tasks at test time. However, when the goal space has discontinuities and the reward is sparse, a majority of goals are difficult to reach. In this context, a curriculum over goals helps agents learn by adapting training tasks to their current capabilities. In this work we propose Stein Variational Goal Generation (\svgg), which samples goals of intermediate difficulty for the agent, by leveraging a learned predictive model of its goal reaching capabilities.
The distribution of goals is modeled with particles that are attracted in areas of appropriate difficulty using Stein Variational Gradient Descent. We show that \svgg outperforms state-of-the-art multi-goal Reinforcement Learning methods in terms of success coverage in hard exploration problems, and demonstrate that it is endowed with a useful recovery property when the environment changes.
\end{abstract}

\section{Introduction}

Multi-goal Reinforcement Learning (RL) \cite{kaelbling1993learning}, where agent policies are conditioned by goals specified according to the task at hand, has recently been at the heart of many research works \cite{schaul2015universal,pitis2020maximum, yang2021density}, since it offers an efficient way for sharing experience between related tasks. The usual ambition in the multi-goal RL context is to obtain an agent able to reach any goal from some desired goal distribution. % and to do so reliably and efficiently. 
This is particularly challenging in settings where the desired goal distribution is unknown at train time, which means discovering the space of valid goals by experience, and optimizing {\em success  coverage}\footnote{{\em Success  coverage},  which measures the average performance of the agent on all valid goals, is denoted {\em competence} in \cite{blaes2019control}.}

%and {\em feasibility} in \cite{racaniere2019automated}.} without any prior knowledge.  

To avoid deceptive gradient issues and a tedious reward engineering process, multi-goal RL often considers the sparse reward context, where the agent only obtains a non-null learning signal when the goal is achieved. In that case, the multi-goal framework makes it possible to leverage Hindsight Experience Replay (\her) \citep{andrychowicz2017hindsight} which helps densify the reward signal by relabeling failures as successes for the goals achieved by accident.
However, in settings with discontinuities in the goal space (e.g., walls in a maze), or in hard exploration problems where the long task horizon results in an exponential decrease of the learning signal \citep{osband2016generalization}, many goals remain hard to achieve and using \her does not suffice to reach all valid goals. 
%In these more difficult contexts, and without any desired goal distribution at hand, we want to maximize a metric that we call {\em success coverage}\footnote{The same metric is called {\em competence} in \cite{blaes2019control} and {\em feasibility} in \cite{racaniere2019automated}.} which measures the performance of the agent on all valid goals. This metric encompasses the capacity of the agent to explore and master every valid goal in the environment.
In these more difficult contexts, and without any desired goal distribution at hand, the selection of training goals from a behavior distribution must be structured into a curriculum to help agents explore and learn progressively by adapting training tasks to their current capabilities \citep{colas2022autotelic}. The question is: how can we organize a curriculum of goals to maximize the success coverage?

A first approach consists in focusing on novelty, with the objective of expanding the set of achieved goals. This is the approach of \rig \cite{nair2018visual}, \discern \cite{warde2018unsupervised}, \skewfit\cite{pong2019skew} and \mega   \cite{pitis2020maximum}\footnote{This is also the case for \omeg \cite{pitis2020maximum}, which extends \mega, but for settings where the desired goal distribution is known, which leaves the scope of our work.}. This leads to strong exploration results, but success coverage is only optimized implicitly.

Another strategy is to bias the goal generation process toward goals of intermediate difficulty (GOIDs). The general intuition is that addressing goals that are too easy or too hard does not foster progress, thus the agent needs to identify goals on which it can make some progress. The focus is thus more on performance. This is the approach of asymetric self-play \cite{sukhbaatar2017intrinsic}, \ggan \cite{florensa2018automatic}, \setso \cite{racaniere2019automated} or \vds \cite{zhang2020}. By aiming at performance, those methods target more explicitly success in encountered goals, but benefit from implicit exploration.

In this work, we propose a novel method which provides the best of both worlds. Our method, called \svgg\footnote{For Stein Variational Goal Generation.}, learns a model of the probability of succeeding in achieving goals by relying on a set of particles where each particle represents a goal candidate. This set of particles is updated via Stein Variational Gradient Descent \citep{svgd} to fit GOIDs, modeled as goals whose success is the most unpredictable. A key feature of \svgd is that the gradient applied to particles combines an attraction term and a repulsion term, which helps strike a good balance between exploration and exploitation. In particular, when the agent cannot find any additional GOID, due to the repulsion term of \svgd, the current particles repel one another resulting in fostering more exploratory goal sampling.
This endows \svgg with a very flexible model of the current capabilities of the corresponding  agent. Based on this feature, we formally demonstrate that \svgg possesses a very useful \textit{recovery property} that prevents catastrophic forgetting and enables the agent to adapt when the environment changes during training. We empirically validate \svgg on Multi-goal RL problems where the goal space is of moderate size, and leave investigations on problems where goals are images for future work.
%and do not compare to algorithms such as \rig \cite{nair2018visual}, \discern \cite{warde2018unsupervised} or \setso \cite{racaniere2019automated} which are dedicated
%\input{intro_olivier.tex}

\section{Background}

\subsection{Goal-conditioned Reinforcement Learning}

%In this paper, we consider a standard multi-goal reinforcement learning setting where agents are conditioned on a {\em behavioral goal} from some goal space, their interaction with the environment results in some {\em achieved goal} distribution, and they may be expected to reach some {\em desired goal} distribution \citep{liu2022goal}. In general, the agent does not know in advance if a behavioral goal is {\em valid}, i.e. whether it can be achieved or not.

%More formally, the framework is 
In this paper, we consider the multi-goal reinforcement learning setting,  defined as a Markov Decision Process (MDP) ${\cal M}_g=<S,T,A, g, R_g>$, where $S$ is a set of states, $T$ is the set of transitions, $A$ the set of actions and the reward function $R_g$ is parametrized by a goal $g$ lying in the d-dimensional continuous goal space ${\cal G} \equiv \mathbb{R}^d$. In our setting, each goal $g$ is defined as a set of states $S_g \subseteq S$ that are desirable situations for the corresponding task, with states in $S_g$ being terminal states of the corresponding MDP. Thus, a goal $g$ is considered achieved when the agent reaches at step $t$ any state $s_t \in S_g$, which implies the following sparse reward function $R_g:S \rightarrow \{0;1\}$ in the absence of expert knowledge. The goal-conditioned reward function is defined as $R_g(s_t,a_t,s_{t+1})=I(s_{t+1} \in S_g)$ for discrete state spaces and $R_g(s_t,a_t,s_{t+1})=I(\min_{s^* \in (S_g)}||s_{t+1} - s^*||_2 < \delta)$ for discrete ones, where $\delta$ is a distance threshold and $I$ the indicator function.

Then, the objective is to learn a goal-conditioned policy (GCP) $\pi:S\times {\cal G} \rightarrow A$ which maximizes the expected cumulative reward from any initial state of the environment, given a goal $g \in {\cal G}$: $\pi^*=\arg\max_\pi \mathbb{E}_{g \sim p_d} \mathbb{E}_{\tau \sim \pi(\tau)} [\sum_{t=0}^{\infty} \gamma^t r^g_t]$, where $r^g_t=R_g(s_t,a_t,s_{t+1})$ stands for the goal-conditioned reward obtained at step $t$ of trajectory $\tau$ using goal $g$, $\gamma$ is a discount factor in $]0;1[$ and $p_d$ is the distribution of goals over ${\cal G}$. In our setting we consider that $p_d$ is uniform over $S$ (i.e., no known desired distribution), while the work could be extended to cover different distributions.   

Importantly, since $S$ is not known in advance and we want to adapt training goals to the current capabilities of the agent, learning is performed at each step through goals sampled from a behavioral distribution $p_{goals}$, which is periodically updated by experience. While training the agent can be performed by any classical RL algorithm, our work focuses on the definition of this behavioral distribution $p_{goals}$, which drives the learning process.

\subsection{Automatic Curriculum for sparse Reward RL}

Our \svgg method addresses automatic curriculum for sparse reward goal-conditioned RL (GCRL) problems and learns to achieve a continuum of related tasks. 

\paragraph{Achieved Goals Distributions} Our work is strongly related to the \mega algorithm \citep{pitis2020maximum}, which (1) maintains a buffer of previously achieved goals, (2) models the distribution of achieved goals via a kernel density estimation (KDE), and (3) uses this distribution to define its behavior distribution. By preferably sampling from the buffer goals at the boundary of the set of already reached states, an increase of the support of that distribution is expected. In that way, \mega aims at overcoming the limitations of previous related approaches which also model the distribution of achieved goals. For instance, \discern \citep{warde2018unsupervised} only uses a replay buffer of goals whereas \rig \citep{nair2018visual} and \skewfit \citep{pong2019skew} rather use variational auto-encoding \citep{kingma2013auto} of the distribution. While \rig samples from the model of the achieved distribution, and \discern and \skewfit skew that distribution to sample more diverse achieved goals, \mega rather focuses on low density regions of the distribution, aiming to expand it. This results in improved exploration compared to competitors. Our approach differs from all these works as they only model achieved goals, independently from which goal was targeted when they were achieved, whereas we model the capability of reaching target goals. This makes a strong difference since, while \mega selects goals at the frontier of what it already discovered, nothing indicates that goals $g$ closer to the mode of the distribution can be achieved when they are targeted. \mega is also prone to catastrophic forgetting and limits exploration to goals present in the replay buffer. %(goals need to be achieved once to serve as behavior goal candidates). % for conditioning the policy).          

\paragraph{Adversarial Goal Generation} Another trend proposes to adversarially learn a goal generator, that produces targets that are at the frontier of the agent's capabilities. In that vein, \ggan \citep{florensa2018automatic} simultaneously learns a discriminator to sort out non GOIDs or generated goals from GOIDs in the buffer of achieved goals, and a generator that aims at producing goals that fool the discriminator. While appealing, the approach is prone to instabilities, with a generator that may usually diverge far from the space of reachable goals. 
\setso \citep{racaniere2019automated} stands as an extension of \ggan, where a goal setter is learned to provide goals of various levels of difficulty w.r.t. a judge network (similar to our success predictor, see next section). The provided goals remain close to already achieved goals, and are diverse enough to avoid mode collapse. This approach however suffers from relying on invertible networks to map from the latent space to the goal space, which severely limits the modeling power, and can reveal problematic for environments with strong discontinuities. %high dimensional goal spaces, e.g. images. 
Asymmetric self-play \citep{sukhbaatar2017intrinsic} is another way to generate goals, with a teacher agent seeking to produce goals that are just beyond the capabilities of the student agent. Both teacher and student learn simultaneously, with an equilibrium of adverse rewards determined on their respective time to go. However, this balance is hard to maintain, and many useful areas are usually missed. %Finally, curiosity-driven exploration \citep{pathak2017curiosity} could be placed in that adversarial category, with an inverse model that aims at learning a compact latent representation space from which it is still possible to predict the dynamics of the environment. The agent is then rewarded for exploring areas whose dynamics are hard to predict. Nevertheless, it requires complex model learning and may lead to hallucinations for the agent which can remain stuck in difficult areas of the search space, by contrast with our simpler model-free setting.   
Our \svgg algorithm also samples GOIDs, but it does so by learning a predictive model of the agent's goal achievement capability and building a sampling distribution that focuses on goals whose achievement is the most unpredictable.
Because it does not call upon an adversarial generative process, \svgg is less prone to instabilities.

\subsection{Stein Variational Gradient Descent}

Our method builds on Stein Variational Gradient Descent (\svgd) \citep{svgd} to approximate the distribution of goals of interest. \svgd is a powerful non-parametric tool for density estimation, when the partition function of the target distribution $p$ to approximate is intractable, which is the case when we do not know the support of the valid goal distribution in the environment. It stands as an efficient alternative to MCMC methods, which are proven to converge to the true distribution $p$ but are usually too slow to be used in complex optimization processes. It also stands as an alternative to variational inference of parametric neural distributions $q$, which are restricted to pre-specified families of distributions (e.g., Gaussian or mixtures of Gaussians) that may not fit target distributions. Instead, it models %the variational distribution 
$q$ as a set of particles $\left\{x_{i}\right\}_{i=1}^{n}$, all belonging to the support of $p$.

The idea behind \svgd is to approximate the target distribution $p$ with $q$ by minimizing their KL-divergence:  %\begin{equation}
%\label{stein_objective}
    $\min_q KL(q||p)$. 
%\end{equation} 
This objective is reached by iterative deterministic transforms as small perturbations of the identity map, on the set of particles: $T(x) = x+\epsilon \phi(x)$, where $\phi$ is a smooth transform function that indicates the direction of the perturbation, while $\epsilon$ is the magnitude.

The authors draw a connection between KL-divergence and the \textit{Stein operator} $\mathcal{A}_p  \phi(x) =\phi (x) \nabla_x \log p(x)^T + \nabla_x \phi(x)$ by showing that 
%\begin{equation}
%\label{stein_0}
 $   \nabla_{\epsilon}KL(q_{[T]}||p)|_{\epsilon = 0} =  -\mathbb{E}_{x\sim q}[\text{trace}(\mathcal{A}_p \phi(x)]$,
%\end{equation}
 where $q_{[T]}$ is the distribution of particles after the transformation $T$. The KL minimization objective %\eqref{stein_objective}
 is thus related to the \textit{Stein Discrepancy}, defined as:
 $   \mathbb{S}(q,p) = \max\limits_{\phi \in \mathcal{F}} \mathbb{E}_{x\sim q}[\text{trace}(\mathcal{A}_p \phi(x)]$\footnote{Note that $\mathbb{S}(q,p) = 0$ only if $q=p$.}. 

Minimizing Stein Discrepancy being intractable as such, \cite{liu2016kernelized} and \cite{chwialkowski2016kernel} introduce the \textit{Kernelized Stein Discrepancy} (KSD) where the idea is to restrict to projections $\phi$ that belong to the unit ball of a reproducing kernel Hilbert space $\mathcal{H}$ (RKHS), for which there is a closed form solution. The KSD is defined as
$\mathbb{S}(q,p) = \max_{\phi \in \mathcal{H}} \{ \mathbb{E}_{x\sim q}[\text{trace}(\mathcal{A}_p \phi(x)], \quad s.t \quad ||\phi||_{\mathcal{H}} \leq 1 \}
$, whose solution is given by:  
$ \phi^*(.) = \mathbb{E}_{x\sim q}[\mathcal{A}_p k(x,.)]$,
where $k(x,x')$ is the positive definite kernel of the RKHS $\mathcal{H}$. The RBF kernel $k(x,x') = \exp (-\frac{1}{h}||x-x'||^2_2)$ is commonly used.

Therefore, the steepest descent on the KL-objective is given by the optimal transform $x_{i} \leftarrow x_{i} + \epsilon \phi^*(x_i), \quad \forall i=1 \cdots n$, where 

\begin{equation}
\begin{split}
\label{optimal_transform}
\phi^*(x_i) = &\frac{1}{n}\sum_{j=1}^n \big [  \underbrace{k(x_j,x_i) \nabla_{x_j} \log p(x_j)}_{\text{attractive force}}\\& + \underbrace{\nabla_{x_j} k(x_j,x_i)}_{\text{repulsive force}} \big ].
\end{split}
\end{equation}

The “attractive force” in the update \ref{optimal_transform} drives the particles toward high density areas of the target $p$. The “repulsive force” pushes the particles away from each other, therefore fosters exploration and avoids mode collapse. Note that if $n=1$, the update in \eqref{optimal_transform} corresponds to a Maximum a Posteriori.

\svgd has already been successfully explored in the context of RL. The Stein Variational Policy Gradient (\svpg) \citep{liu2017stein} employs \svgd to maintain a distribution of efficient agents as particles. It strongly differs from our approach, since we consider particles as behavior goal candidates, while \svpg aims at capturing the epistemic uncertainty about policy parameters. 
\cite{https://doi.org/10.48550/arxiv.2111.04613} also relies on \svgd to build a strategy to generate goals to agents, but in a very simplified setting without the attractive force from \eqref{optimal_transform}, which prevents from fully benefiting from this theoretical framework. Notably, such a kind of approach is particularly sensitive to catastrophic forgetting.

\section{Stein Variational Goal Generation}

\begin{figure*}
    \centering
    \includegraphics[width=0.9\textwidth]{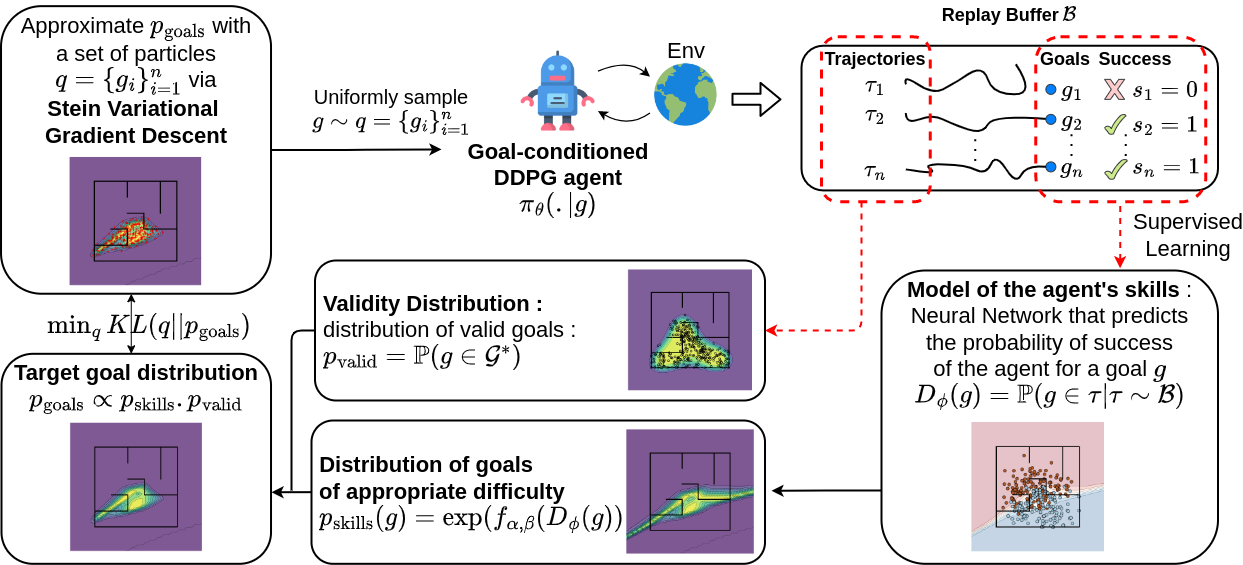}
    \caption{Overview of the \svgg method. The interaction of agents with their environment is stored in a replay buffer (top right) and used to learn $D_{\phi}$, a model of its abilities to achieve goals (bottom right). We build on this model to compute a distribution of goals of appropriate difficulty $p_{\text{skills}}$, leveraging a validity distribution $p_{\text{valid}}$ to stay inside the space of valid goals. The obtained behavioral goal distribution $p_{\text{goals}}$ is approximated with particles $\{g_i\}_{i=1}^n$ using Stein Variational Gradient Descent (\svgd), and the agent samples a goal from these particles.}
    \label{fig:img1}
\end{figure*}

\begin{algorithm*}[ht]
\caption{\textit{RL with Stein Variational Goal Generation}}
\label{algo:svgg_court}
\begin{algorithmic}[1]
\STATE \textbf{Input:} a GCP $\pi_\theta$, % with parameters $\theta$,
%a parameterizable environment ${\cal M}$, 
a success predictor $D_\phi$, % with parameters $\phi$,
a reachability predictor $V_\psi$, 
buffers of transitions ${\cal B}$, reached states ${\cal R}$ and success outcomes $O$, a kernel $k$. %, numbers $t^{(r)}$, $t^{(m)}$, $t^{(p)}$, $l^{(r)}$, $l^{(m)}$ and $l^{(p)}$.  
%\STATE Set $p_c$ as the uniform distribution on $C$;
\STATE Sample a set of particles $q=\{x^i\}_{i=1}^m$ uniformly from states reached by pre-runs of $\pi_\theta$; % in a sequence of pre-runs using random goals; 
\FOR{$n$ epochs}  %\\[0.2cm]
%$\qquad \qquad \qquad \qquad \qquad \qquad \qquad \qquad \qquad \quad$$\vartriangleright$ \textit{Model Update} 
%\STATE blabla \COMMENT{\textit{Data Collection}} 
%\Statex{\hspace{1cm} \textit{\#\#\#\#\#\#\# Data Collection \#\#\#\#\#\#\#}}
\STATE $\vartriangleright$\textit{ Data Collection:} Perform Rollouts of $\pi_\theta$ conditioned on goals uniformly sampled from $q$;
\STATE $\qquad$ Store transitions in ${\cal B}$, visited states in $\cal R$ and success outcomes in ${\cal O}$; %$ Sample 
\label{store1A_court} 
%\ENDFOR 
%\\[0.2cm] 
%\Statex {\hspace{1cm} \textit{\#\#\#\#\#\#\# Model Update \#\#\#\#\#\#\#}}
%\FOR[$\qquad \qquad \qquad \qquad \qquad \qquad \qquad \qquad \qquad \quad$
\STATE $\vartriangleright$ \textit{Model Update}
%\STATE $\qquad$ Sample a batch %of $l^{(m)}$ outcomes \\ $\qquad 
%$\{(g^i,s^i)\}_{i=1}^{l^{(m)}}$ %uniformly
%from $O$; 
\STATE $\qquad$ Update model $D_\phi$ (e.g., via ADAM), using gradient of ${\cal L}_\phi$ \eqref{model_loss} with samples from ${\cal O}$; 
%\ENDFOR %\\[0.2cm] 
%\Statex {\hspace{1cm} \textit{\#\#\#\#\#\#\# Prior  Update \#\#\#\#\#\#\#}}
\STATE $\vartriangleright$ \textit{Prior  Update}
\STATE $\qquad$ Update model $R_\psi$ according to all states in ${\cal R}$; % $\qquad \qquad \quad \qquad \qquad$$\vartriangleright$ \textit{Prior  Update}
%\Statex {\hspace{1cm} \textit{\#\#\#\#\#\#\# Particles Update \#\#\#\#\#\#\#}}
%\STATE Compute $K$ as the Gram matrix  of particles in $\Omega$ using kernel $k(.,.)$;
%\STATE Compute tensor $R=[\nabla_{x_i} k(x_i,x_j)]_{(x_i,x_j) \in \Omega^2}$;
%\STATE Compute $D=[\log(\dfrac{D_\phi(x_i)}{1-D_\phi(x_i)}) \nabla_{x_i}D_\phi(x_i)]_{x_i \in \Omega}$;
\STATE $\vartriangleright$ \textit{Particles Update} (t steps) 
%\FOR{ some iterations} 
%\STATE Compute the density of the target $p_{\text{goals}}$ for the set of particles $\Omega$ using \ref{SVGG_distrib}; 
%\STATE $\{{p_{\text{goals}}(x_i)}\}_{i=1}^n = \{{p_{\text{skills}}(x_i) . p_{\text{prior}}(x_i)}\}_{i=1}^n$
%\STATE Compute transforms: % for every particle $i$ from $\Omega$:  %$\{\phi^*(x_i)\}_{i=1}^n$
%\STATE
%$\phi^*(x_i) =\frac{1}{m}\sum_{j=1}^m \big [  k(x_j,x_i) \nabla_{x_j} \log p_{\text{goals}}(x_j) + \nabla_{x_j} k(x_j,x_i) \big ]$; %equivalent to \eqref{optimal_transform}
\STATE $\qquad$ Update particles $x_{i} \leftarrow x_{i} + \epsilon \frac{1}{m}\sum_{j=1}^m \big [  k(x_j,x_i) \nabla_{x_j} \log p_{\text{goals}}(x_j) + \nabla_{x_j} k(x_j,x_i) \big ]$;
%\phi^*(x_i), \quad \forall i=1 \cdots n$;
%\STATE Sample a batch $\{x^i\}_{i=1}^{l^{(p)}}$ from $\Omega$;
%\STATE Update particles from the batch using $K$, $R$ and $D$ , equivalent to \eqref{particle_moves2} ; 
%\ENDFOR
%\\[0.2cm]
\STATE $\vartriangleright$
 \textit{Agent Improvement}
\STATE $\qquad $ Improve agent with any Off-Policy RL algorithm   (e.g., \ddpg) using transitions from $\cal B$;
 %\Statex{\hspace{1cm} (e.g., \ddpg) using transitions in $\cal B$;}
\ENDFOR
%\Return{ } %$\pi_\theta$
\end{algorithmic}
\end{algorithm*}

In this section we introduce our Stein Variational Goal Generation (\svgg) algorithm. The pseudo-code of \svgg is given in Algorithm~\ref{algo:svgg_court} (a more detailed version is given in  Appendix~\ref{sec:svgg_algo}). Our aim is to obtain a curriculum to sample goals of appropriate difficulty for the RL agent, where the curriculum is represented as an evolving goal sampling probability $p_{\text{goals}}$. To do so, we maintain a model of the agent’s skills – or goal reaching capability –, which helps define a distribution $p_{\text{skills}}$. This distribution assigns probability mass to areas of goals of appropriate difficulty. 
Additionally, with a simple one class SVM, we learn a validity distribution $p_{\text{valid}}$ preventing the particles from being sampled from non-valid areas of the goal space. Then, we aim at sampling training goals from the following target distribution:

\begin{equation}
\label{SVGG_distrib}
    p_{\text{goals}}(g) \propto p_{\text{skills}}(g) . p_{\text{valid}}(g).
\end{equation}

Since computing the partition function is intractable for such a distribution formulation, we rather sample uniformly over a set of particles $q=\left\{x_{i}\right\}_{i=1}^{m}$, that are optimized through SVGD to approximate  $p_{\text{goals}}(g)$. Importantly, for our setting where we are interested in tracking useful areas to train the agent, dealing with a set of particles representing the state of the full exploration landscape appears better fitted than methods relying on single candidates, such as MCMC or Langevin Dynamics, that would produce samples very correlated in time, with unstable dynamics.
This choice also improves the interpretability of the process, by providing a comprehensive picture of the current behavior distribution along training.
Formal definitions of the two components of $p_{\text{goals}}$ are given below.

\paragraph{Model of the agent's skills} The probability $p_{\text{skills}}$ is modeled as a Neural Network $D_{\phi}$ whose parameters $\phi$ are learned by gradient descent on the following Binary Cross Entropy (BCE) loss: 

\begin{equation}
\label{model_loss}
{\cal L}_\phi =  \sum_{(g^i,s^i) \in O} s^i (\log D_{\phi}(g^i)) + (1-s^i) (\log (1-D_{\phi}(g^i))),
\end{equation}
where $O = \{ g^i,s^i \}_{i=1}^{n_{B}}$ is a batch of $(goal, success)$ pairs coming from recent trajectories of the agent in the environment. 
The sampled goals are those whose predicted probability of success is neither too high nor too low (i.e., we avoid $D_{\phi}(g) \approx 1$ or $D_{\phi}(g) \approx 0$).

%\begin{wrapfigure}[15]{r}{0.4\textwidth}
    %\centering
    %\includegraphics[width=0.4\textwidth]{figures/betas.png}
  %  \caption{Modulation of goal's target difficulty with beta distributions.}
 %   \label{fig:betas}
%\end{wrapfigure}
To build $p_{\text{skills}}$ based on the prediction of $D_{\phi}$, we use a beta distribution whose maximum density point mass is determined according to the output of $D_\phi$, by two hyper-parameters $\alpha$ and $\beta$ that shape the distribution and control the difficulty of the addressed goals, as illustrated on \figurename~\ref{fig:betas} in appendix.

We define the distribution $p_{\text{skills}}$ as an energy-based density whose potential is the output of the beta distribution $f$:
\begin{equation}
\label{energy_based_distrib}
p_{\text{skills}}(g) \propto \exp{(f_{\alpha, \beta}(D_\phi(g))}.
\end{equation}

In Appendix~\ref{app:alpha_beta}, we compare the relative performance of 5 pairs of $\alpha$ and $\beta$ and show that target a \textit{Medium} difficulty works best. We stick to this setting in the rest of the study.

\paragraph{Validity distribution} As outlined in \cite{racaniere2019automated}, we would like to only sample valid goals. To do so, instead of their {\em validity loss}, we define a validity distribution which represents the probability that a goal $g$ belongs to the set of valid (reachable) goals $\mathcal{G}^* \subseteq \mathcal{G}$. However, $\mathcal{G}^*$ is not known in advance.
To circumvent this difficulty, the states already reached by the agent are stored in an archive ${\cal R}$ and we aim at %approximate
defining the validity distribution as depending on the  posterior probability given ${\cal R}$: $p_{\text{valid}}(g) \propto  \mathbb{P}(g \in \mathcal{G}^* |{\cal R} )$. We progressively build this distribution with a One Class SVM (OCSVM). This model is mainly designed for outlier or novelty detection in absence of labeled data. Given a dataset $X \in \mathbb{R}^d$, it defines a boundary of the data support in $\mathbb{R}^d$, while keeping a small portion of the data points out of that boundary. With data points being goals from $\cal{R}$, we get
\begin{equation}
\label{prior_ocsvm}
p_{\text{valid}}(g) \propto % \mathbb{P}(g \in \tilde{\mathcal{G}}|{\cal R})=% \text{supp}(G_{\text{as}}}))=
V_{\psi}(g),
\end{equation}
where $V_{\psi}(g)$ is the output of the OCSVM model trained on $\cal{R}$, with parameters $\psi$. %, and $\tilde{\mathcal{G}}$ is the set of states reached by the agent until the current iteration. 
As the agent progresses and expands its set of achieved states through training, it eventually reaches the environment boundary. In this case, we can expect $V_{\psi}(g) \approx \mathbb{P}(g \in \mathcal{G}^*|g \in \omega)$ for any area $\omega \subseteq {\mathcal G}$.   %$p_{\text{valid}}(g) \propto \mathbb{P}(g \in \tilde{\mathcal{G}}|{\cal R}) % \mathbb{P}(g \in \mathcal{G}|\mathcal{G}_{as})= %\text{supp}(\mathcal{G_{\text{as}}})) = 
%\approx \mathbb{P}(g \in \mathcal{G})$.

\paragraph{Recovery property}

As demonstrated in Theorem~\ref{th:recov} and empirically validated in Section~\ref{sec:recov_results}, \svgg benefits from a useful recovery property: when the environment suddenly changes, the \svgg agent will spontaneously resample goals in the areas that are affected by the change.

\begin{theorem}
\label{th:recov}
\textbf{Recovery property}: %Assume that particles are not allowed to leave a compact set ${\cal C}$, with ${\cal G} \subseteq {\cal C}$. 
%Let us denote as ${\cal C}$ the hyper-cube containing all goals $g$ such that $V_\psi(g)>0$.
Let us denote as ${\cal G}^{+}$ the set of goals $g$ such that $V_\psi(g)>0$ and ${\cal C}\in \mathbb{R}^d$ its convex hull.  %the hyper-cube containing all goals $g\in $.  % $D_\phi(g) \rightarrow 1$. 
Assume that, at a given iteration $l$, $D_\phi(g) \approx 1$ for every $g \in {\cal G}^+$ (i.e., the whole set ${\cal G}^+$ is considered as mastered by $D_\phi(g)$), and that, on that set, $V_\psi$ is well calibrated: for any area $\omega \subseteq {\cal G}^+$ and any goal $g \in \omega$, $V_\psi(g) \approx P(g \in {\cal G}^*|g \in \omega)$. %$V_\psi(g) \approx P(g \in {\cal G})$.
Assume also that we use a kernel which ensures that the Kernelized Stein Discrepancy $KSD(q,p_{\text{goals}})$ of any distribution $q$ with $p_{\text{goals}}$ is 0 only if $q$ weakly converges to $p_{\text{goals}}$\footnote{As assumed for instance in \cite{NIPS2017_17ed8abe}. This is not always true, but gives strong insights about the behavior of \svgd. Refer to \cite{gorham2017measuring} for more discussions about KSD.}. %, and that this divergence is 0 at iteration $l$. %  
Then, with no updates of the models after iteration $l$ and a number of particles $m>1$, %and a sufficiently high number of particles,
any area $\omega \subseteq {\cal G}^+ \cap {\cal G}^*$  with diameter  $diam(\omega)\geq \sqrt{d} \frac{diam({\cal C})}{(\sqrt[d]{m}-1)}$ %, such that $\forall g \in \omega, V_\psi(g) \approx 1$,    
eventually contains at least one particle, whenever $KSD(\{x^i\}_{i=1}^n,p_{\text{goals}})=0$ after convergence of the particle updates.  
\end{theorem}

The above theorem (proof in Appendix \ref{theorem}) ensures that, even if the success model overestimates the capacities of the agent for some area $\omega$ (e.g., due to changes in the environment, catastrophic forgetting or success model error), some  particles are likely to go back to this area once every goal in ${\cal G}^*$ looks well covered by the agent, % (from the $D_\phi$ point of view), 
with an increasing probability for more particles. This way, the process can reconsider overestimated areas, by sampling again goals in them, and hence correcting the corresponding predictions, which leads to attracting attention of $p_{\text{skills}}$ back to these difficult areas. Approaches such as \mega do not exhibit such recovery properties, since they always sample at the boundary of their achievable goal distribution, % $\tilde{\cal G}$,
which is likely to incrementally grow towards ${\cal G}^*$. The particles of our approach can be seen as attention trackers which remodel the behavior distribution and mobilize the effort on useful areas when needed. This is much better than uniformly sampling from the whole space of achievable states with small probability, which would also ensure some recovery of forgotten areas but in a very inefficient way. This theoretical guarantee of \svgg is empirically validated by the experiment from \figurename~\ref{fig:change_maze}, which shows the good recovery property of our approach, after a sudden change in the environment.

\section{Experiments}
\label{xp}

\subsection{Experimental setup}

\paragraph{Success coverage metric}

Our general objective is to obtain a policy that can reliably achieve all valid goals in some environment. To quantify this objective, we evaluate the resulting policy on the entire space of valid goals $\mathcal{G}^*$ in our environment using a success coverage metric, defined as $S(\pi) = \frac{1}{{\cal V}(\mathcal{G}^*)} \int_{\mathcal{G}^*} \mathbb{P}(\pi \text{ achieves } g)dg$, with ${\cal V}(\mathcal{G}^*)$ the volume of $\mathcal{G}^*$. The goal space being continuous, we evaluate the policy on a finite  subset $\hat{\mathcal{G}}$ uniformly sampled from $ \mathcal{G}^*$. Then our objective reduces to:
\begin{equation}
\label{success_coverage_1}
\begin{split}
    S(\pi) %&= \frac{1}{|\hat{\mathcal{G}|}} \sum_{i=1}^{|\hat{\mathcal{G}}|} \mathbb{P}(\pi \text{ achieves } g_i)\\ 
    &=\frac{1}{|\hat{\mathcal{G}|}} \sum_{i=1}^{|\hat{\mathcal{G}}|} \mathbb{E}_{\tau \sim \pi}[\mathbb{1}\{\exists s \in \tau, \min\limits_{s^* \in {\cal S}_{g_i}} ||s - s^*||_2 < \delta\}].
\end{split}
\end{equation}
To build $\hat{\mathcal{G}}$, we split $\mathcal{G}^*$ into areas following a regular grid, and then uniformly sample 30 goals inside each part of the division.

\paragraph{Compared Approaches}
\label{sec:baselines}

As baselines, we choose two methods from the literature that span the existing trends in unsupervised RL with goal-conditioned policies, \mega \cite{pitis2020maximum} and \ggan \cite{florensa2018automatic}. In addition, the Random baseline randomly selects the behavior goal among past achieved states. 

We also evaluate two \svgg ablations: a \textit{No Validity Distribution} version, which considers $p_{\text{goals}} \propto p_{\text{skills}}$ and a \textit{Only Validity} version,  where $p_{\text{goals}} \propto p_{\text{valid}}$. 

Learning to reliably achieve all valid goals can be decomposed in learning to sample appropriate behavioral goals, and learning to achieve these goals. Our focus being on the goal sampling mechanism, all compared approaches learn to achieve goals using \ddpg and \her with a mixed strategy described in Appendix~\ref{app:methods}. In all versions of \svgg, we use $m=100$ particles to approximate the target distribution. Implementation details about all considered architectures are given in Appendix~\ref{app:methods}.

\paragraph{Questions}

To compare the performance of \svgg to baselines, we investigate the following questions:
1) Does \svgg maximize the success coverage metric better than the baselines? 
3) Is \svgg more robust than the baselines to catastrophic forgetting that may occur in sudden environment changes?
In Appendix~\ref{app:alpha_beta}, we also investigate the impact of target difficulty (i.e., beta distribution as described above) on success coverage maximization.

\paragraph{Evaluation environments}

To answer the above metric-oriented questions, we use a modified version of the FetchReach and FetchPush environments \cite{1802.09464} where we have added obstacles in the workspace of the robot arm to increase the amount of discontinuities between the optimal goal-oriented behaviors. We also compare \svgg to baselines in the U-shaped AntMaze environment \cite{https://doi.org/10.48550/arxiv.1911.01417} and in a hard version of FetchPickAndPlace.
Additionally, to provide more analysis-oriented visualizations, we use a Point-Maze environment where an agent moves a point without mass within a 2D maze with continuous spaces of states and actions. As the agent does not perceive the walls, maximizing success coverage in these environments is harder than it seems.

\begin{figure*}[!ht]
    \centering
    %\vspace{-1cm}
        \includegraphics[width=1\textwidth]{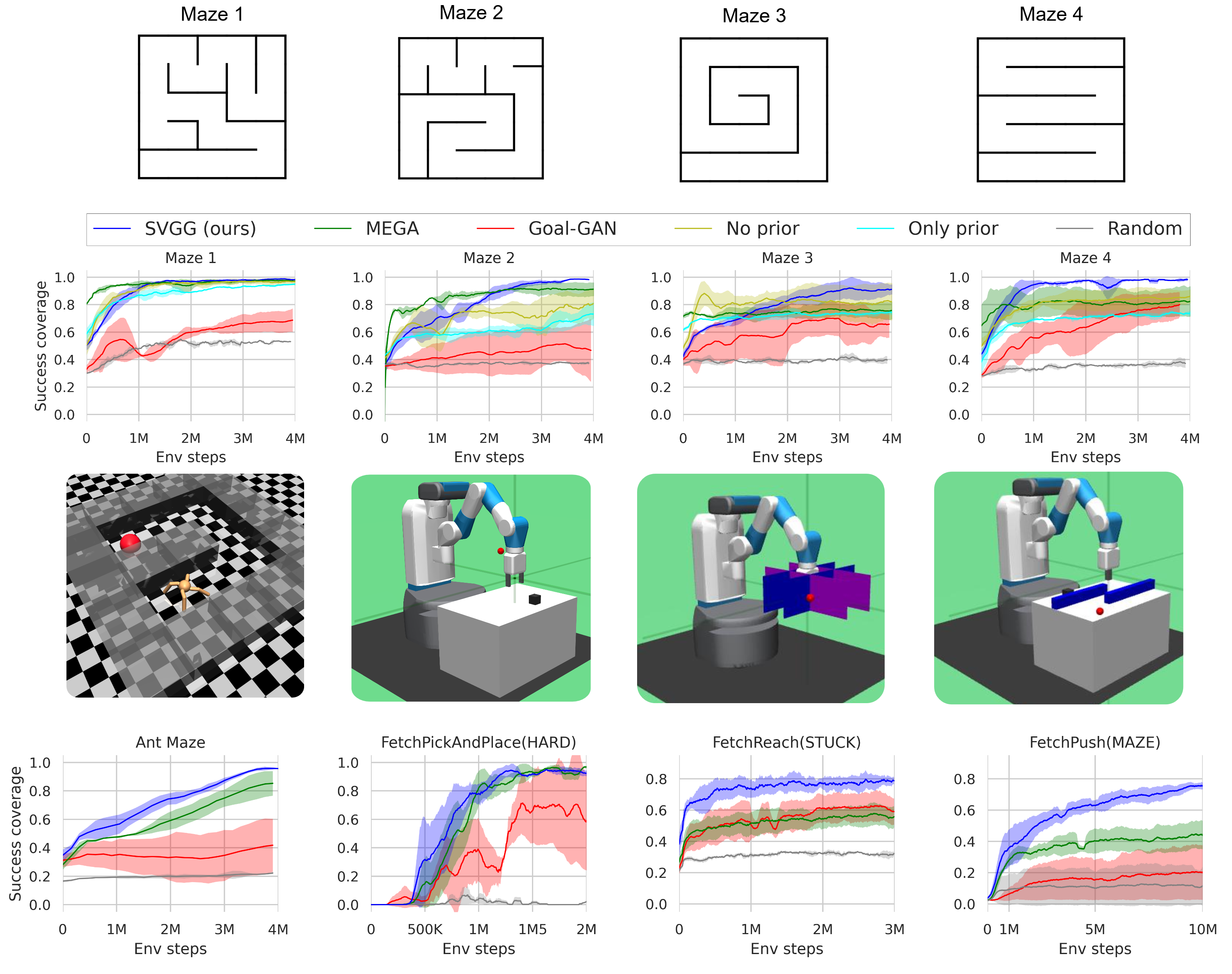}
    \caption{\label{fig:main} Success coverage over 4 different PointMazes, AntMaze, FetchPickAndPlace (Hard), FetchReach (Stuck) and FetchPush (Maze) (4 seeds each). \svgg outperforms \mega and \ggan as well as ablations.}
\end{figure*}

\subsection{Main results}

\paragraph{Success coverage evaluations}

%\figurename~\ref{fig:main} shows the success coverage of baselines over 4 different maze environments. \svgg is the only method that discovers and achieves all valid goals in 4M agent steps in all considered environments. 

\figurename~\ref{fig:main} shows that \svgg significantly outperforms all baselines in terms of success coverage. Especially in highly discontinuous goal space settings such as in mazes and the modified version on FetchReach and FetchPush, where it efficiently discovers and achieves most of the valid goals. On AntMaze and FetchPickAndPlace, where the goal space is more smooth, our approach obtains comparable results to its competitors.

Due to the known stability issues in GAN training, \ggan is the least efficient baselines. Another explanation of the failure of \ggan is that it is likely to generate non valid goals, which is not the case for \mega or \svgg. \mega chooses behavior goals from a Replay Buffer of achieved goals, while the validity distribution $p_{\text{valid}}$ considered in \svgg keeps particles inside valid areas. 

%Nevertheless, we observe that \ggan performs better in environments with linear structures, such as Mazes 3, 4 and FetchPickAndPlace. On the other hand, it has a hard time to generate goals of intermediate difficulty when the structure is random, as in Mazes 1, 2.

The minimum density heuristic of \mega efficiently discovers all valid goals in the environment, but our results show that its success plateaus in almost all the considered environments. \mega's intrinsic motivation only relies on state novelty. Thus, when the agent has discovered all reachable states, it is unable to target areas that it has reached in the past but has not mastered yet. 

%\figurename~\ref{fig:main} also reports results on AntMaze and a hard version of FetchPickAndPlace. \svgg also outperforms \mega on AntMaze, because walls still induce discontinuities. On FetchPickAndPlace, where the goal space is more smooth, our approach obtains comparable results to its competitors. 

\begin{figure*}[!ht]
    \centering
    %\vspace{-0.5cm}
    \includegraphics[width=1\textwidth]{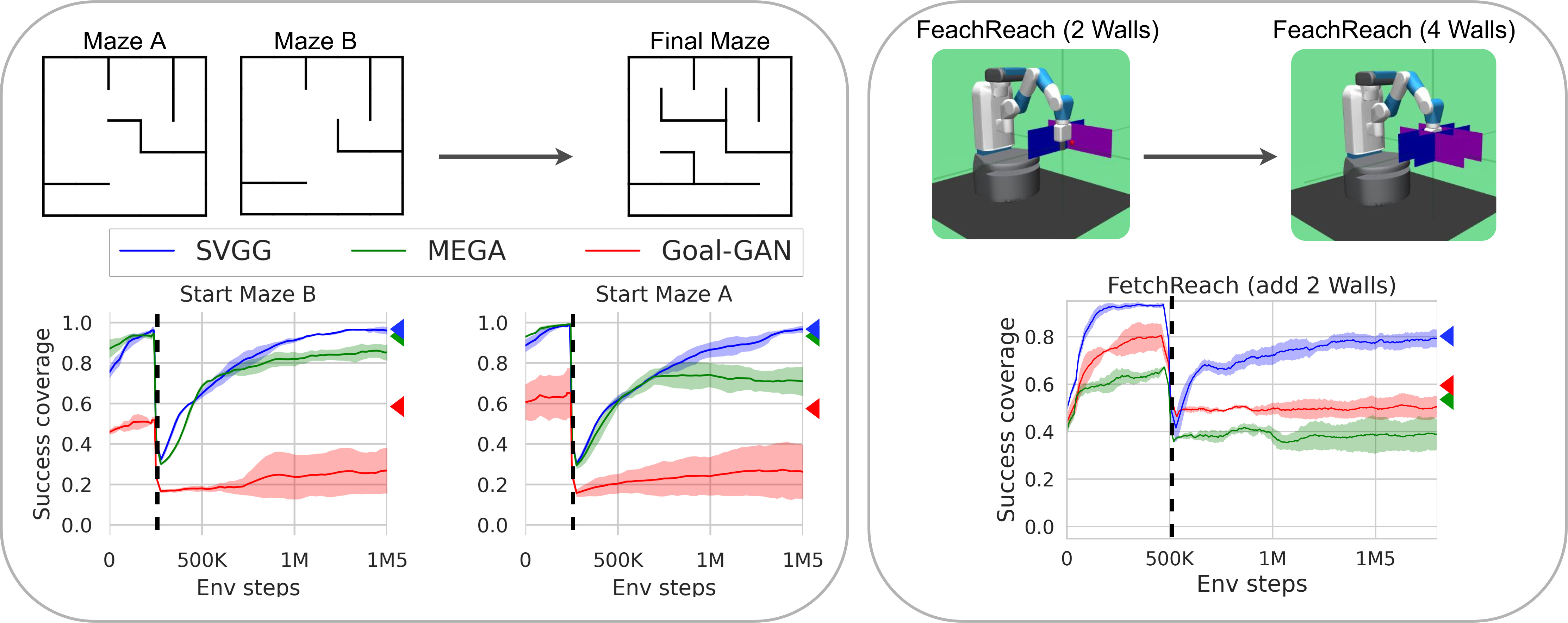}
    \caption{Evolution of success coverage in a changing environment for \mega, \ggan and \svgg (4 seeds each). We add walls to the starting mazes A and B (left) and go from 2 to 4 walls in FetchReach (right).
    The triangles correspond to the pick success coverage of the methods on the final environments (which correspond to maze 1 and FetchReach (Stuck) \figurename~\ref{fig:main}) during regular training.}
    \label{fig:change_maze} 
    %\end{figure}
\end{figure*}

\paragraph{Recovery Property}
\label{sec:recov_results}
\figurename~\ref{fig:change_maze} shows the advantages of \svgg over \mega and \ggan in changing environments. 
Walls are suddenly added during the training process (dot black line from \figurename~\ref{fig:change_maze}), after the methods had reached their pick performance. We see that the performance of \mega and \ggan plateaus lower than their pick performance (see \figurename~\ref{fig:main}) whereas \svgg discovers new difficulties resulting from the environment modification and focuses on those to finally recover its pick success coverage.

%Curves show that \mega is unable to fully adapt to the new % configuration of the 
%maze setting, which corresponds to  Maze 1 from \figurename~\ref{fig:img1} (although \mega solved Maze 1 in 1M steps in that experiment with constant conditions). %when it is fixed.

%On the other hand, \svgg discovers new difficulties resulting from the environment modification and focuses on those to finally solve the entire maze in less than 1.5M steps.

We also observe that the advantages of our method over \mega in terms of recovery ability are more significant when changes in the environment are more drastic (i.e., when starting from maze B).

\begin{figure*}[!ht]
    \centering
    \includegraphics[width=1\textwidth]{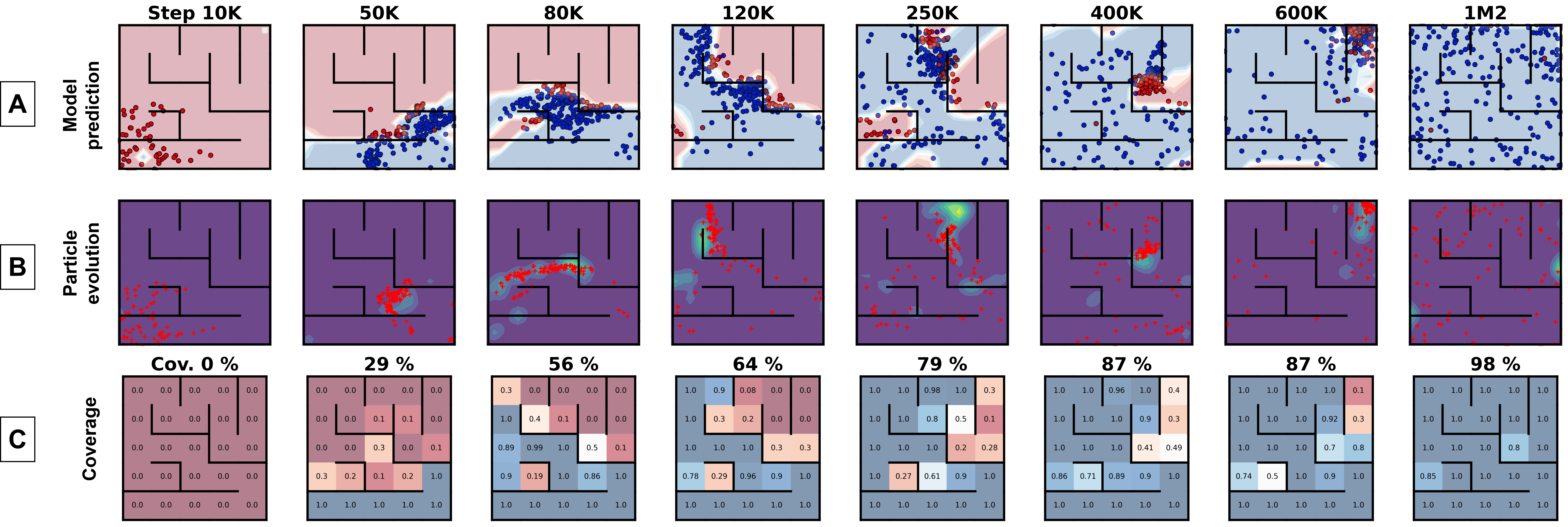}
    \caption{Parallel visualization of the model's prediction (A), the particles evolution (B) and of the coverage (C) throughout training in Maze 1. }%The particles approximate the target difficulty $p_{\text{goals}}$ until the models detects that the agent has nothing else to learn.}
    \label{fig:particles_evo}
\end{figure*}

\paragraph{Further SVGG analyses}

To gain further intuition on how \svgg maximizes the success coverage, we show in \figurename~\ref{fig:particles_evo} the evolution of the particles throughout training. As the agent progresses and achieves novel and harder goals, the $D_{\phi}$ model updates its predictions. Thus, the target distribution $p_{\text{goals}}$ is updated accordingly (background color of the $2^{nd}$ row of the figure). The particles $q=\{g_i\}_{i=1}^n$ are then moved toward new areas of intermediate difficulty through \svgd  to minimize $KL(q||p_{\text{goals}})$.

\figurename~\ref{fig:particles_evo} also highlights the recovery property of \svgg. When the agent has nothing else to learn in the environment, $p_{\text{goals}}$ reduces to $p_{\text{valid}}$, that is at this point uniform over the entire goal space. Therefore, the particles spread uniformly over the goal space and prevent \svgg from catastrophic forgetting, as the model rediscovers areas that the agent has forgotten how to reach (cf. rightmost column in \figurename~\ref{fig:particles_evo}). Additional analyses on \svgg are given in Appendix~\ref{app:alpha_beta}.

%%%%%%%%%%%%%%%%%%%%%%%%%%%%%%%%%%%%%%%%%%%%%%%%%%%%%%%%%%%%%%%%%%%
\section{Conclusion}
\label{conclusion}

This paper introduces a new multi-goal reinforcement learning algorithm, \svgg, which leverages Stein Variational Gradient Descent to monitor a model w.r.t. its goal achievement capabilities. Using this model, the agent addresses goals of intermediate difficulty, resulting in an efficient curriculum for finally covering the whole goal space. Moreover, \svgg can recover from catastrophic forgetting, which is a classic pitfall in multi-goal RL.

Studying the impact of the number of particles is left for future work. Actually, the target distribution being in constant evolution, the KL divergence minimization objective is hard to reach at all times, which makes it difficult to claim that using more particles is always better. Furthermore, a previous work \cite{DBLP:journals/corr/abs-2101-09815} spotted exploration failures in \svgd, and suggests that periodically annealing the attraction force in particle optimization \eqref{optimal_transform} is required to enable particles to cover non-trivial distributions, e.g. in multimodal settings (which is the case for us) or in high dimensions.

Some limitations should be addressed in future work. The environments used in our experiments have low dimensional goal space, which facilitates the approximation of the target distribution with \svgd and the agent's model learning phase. When the agent's observation and goal space will be images, the agent should learn a compact latent space of observation as in \cite{pong2019skew, https://doi.org/10.48550/arxiv.2102.11271, https://doi.org/10.48550/arxiv.2103.04551, pmlr-v97-hafner19a}, using various representation learning techniques like contrastive learning, prototypical representations or variational auto-encoders. In future work, we should learn a latent goal space from observations, and perform \svgd over particles in this latent space. This would result in an end-to-end algorithm learning to discover and achieve all possible goals in its environment from pixels.

Besides, we also envision to address larger and stochastic environments, where additional uncertainty estimation should be added to the goal generation process, to prevent the agent getting stuck in uncontrollable states (like a TV screen showing white noise) as in \cite{https://doi.org/10.48550/arxiv.1805.12114, https://doi.org/10.48550/arxiv.1906.04161}, using methods such as model disagreement between multiple agents.

\section*{Acknowledgments} We thank our colleges at ISIR and the anonymous reviewers for their helpful comments. This work benefited from the use SCAI (Sorbonne Center for Artificial Intelligence) cluster of GPUs. This work has received funding from the European Commission's Horizon Europe Framework Programme under grant agreement No 101070381 (PILLAR-robots project).

\bibliography{biblio}
\bibliographystyle{icml2023}

\newpage

\appendix
\section{Proof of Theorem 1}
\label{theorem}

%\begin{theorem}
\begin{customthm}{1}
\textbf{Recovery property}: %Assume that particles are not allowed to leave a compact set ${\cal C}$, with ${\cal G} \subseteq {\cal C}$. 
%Let us denote as ${\cal C}$ the hyper-cube containing all goals $g$ such that $V_\psi(g)>0$.
Let us denote as ${\cal G}^{+}$ the set of goals $g$ such that $V_\psi(g)>0$ and ${\cal C}\in \mathbb{R}^d$ its convex hull.  %the hyper-cube containing all goals $g\in $.  % $D_\phi(g) \rightarrow 1$. 
Assume that, at a given iteration $l$, $D_\phi(g) \approx 1$ for every $g \in {\cal G}^+$ (i.e., the whole set ${\cal G}^+$ is considered as mastered by $D_\phi(g)$), and that, on that set, $V_\psi$ is well calibrated: for any area $\omega \subseteq {\cal G}^+$ and any goal $g \in \omega$, $V_\psi(g) \approx P(g \in {\cal G}^*|g \in \omega)$. %|g \in {\cal G}^+)$.  %% for any $g \in ${\cal G}^+${\tilde{\cal G}} \subseteq {\cal G}^+$ % everywhere  $V_\psi(g) \rightarrow 1$  
Assume also that we use a kernel which ensures that the Kernelized Stein Discrepancy $KSD(q, p_{\text{goals}})$ of any distribution $q$ with $p_{\text{goals}}$ is 0 only if $q$ weakly converges to $p_{\text{goals}}$\footnote{As assumed for instance in \cite{NIPS2017_17ed8abe}. This is not always true, but gives strong insights about the behavior of \svgd. Please refer to \cite{gorham2017measuring} for more discussions about KSD.}. %, and that this divergence is 0 at iteration $l$. %  
Then, with no updates of the models after iteration $l$ and a number of particles $m>1$, %and a sufficiently high number of particles,
any area $\omega \subseteq {\cal G}^+ \cap {\cal G}^*$  with diameter  $diam(\omega)\geq \sqrt{d} \frac{diam({\cal C})}{(\sqrt[d]{m}-1)}$ %, such that $\forall g \in \omega, V_\psi(g) \approx 1$,    
eventually contains at least one particle, whenever $KSD(\{x^i\}_{i=1}^n,p_{\text{goals}})=0$ after convergence of the particle updates.     
\end{customthm}

\begin{figure*}[ht]
    \centering
        \includegraphics[width=0.99\textwidth]{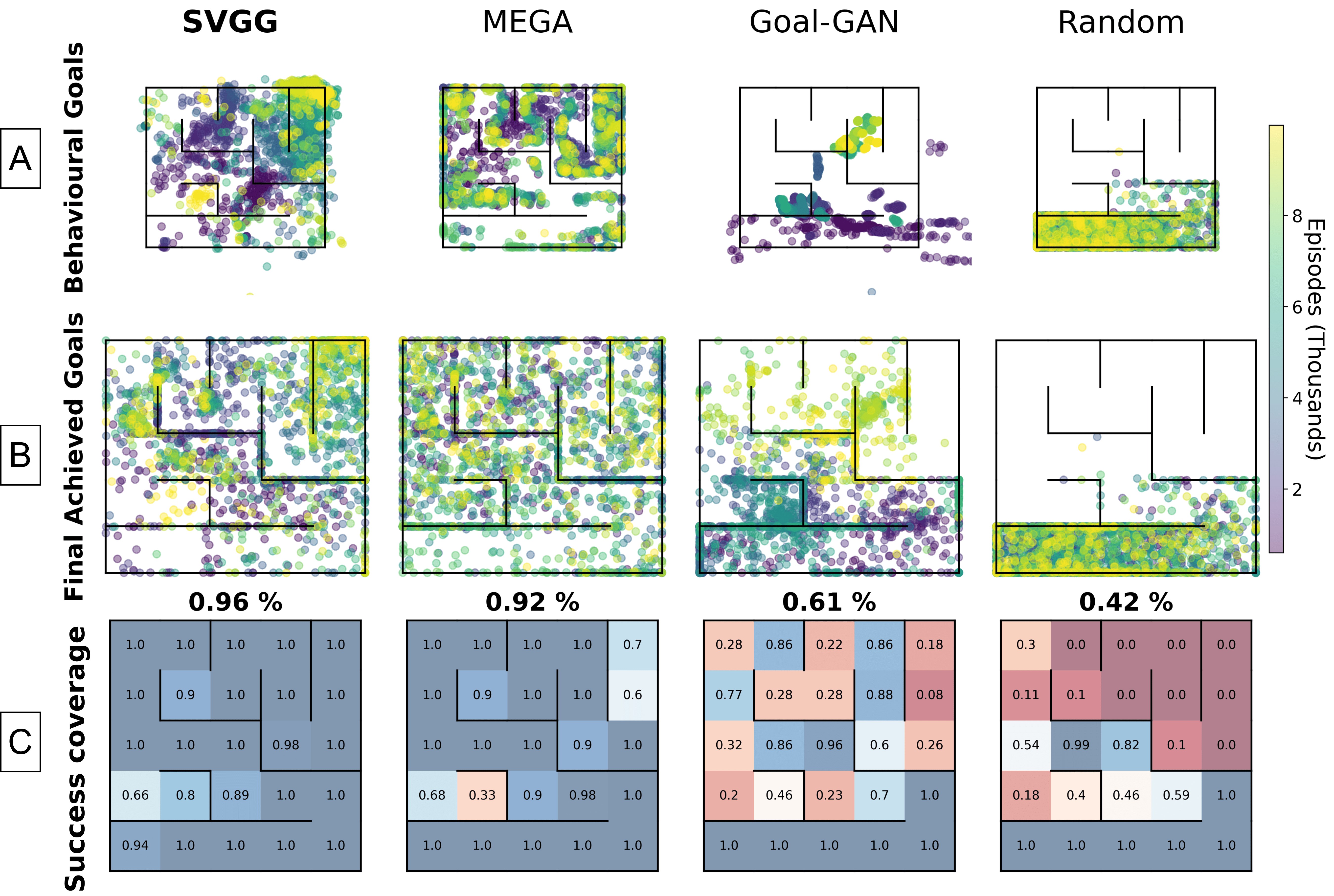}
    \caption{ \label{fig:visu_1}Visualization of: (A) behavioral goals , (B) achieved goals and (C) success coverage for 1M steps in Maze 1.}
\end{figure*}

\begin{proof}

In settings where the KSD of a distribution $\mu$ from a target $p$ is 0 only if $\mu$ weakly converges to $p$, \cite{NIPS2017_17ed8abe} previously proved that, for any compact set, the empirical measure $\mu^n$ of $\mu$, computed on a set of $n$ particles, converges weakly towards the target $p$ when a sufficient number of particles is used.   %after a sufficient number of particles transforms. %$n \rightarrow +\infty$. %This allows to restrict the proof to the convergence of $\mu_l$ towards 
Thus, under our hypotheses, the set of particles of \svgg appropriately represents $p_{\text{goals}}$ after a sufficient number of steps of Stein transforms.

Now, we know that particles cannot leave ${\cal G}^+$ since $V_\psi=0$ outside, and so does $p_{\text{goals}}$. Since  $V_\psi$ is well calibrated on ${\cal G}^+$, we also know that $V_\psi=1$ on every area of ${\cal G}^+$ only containing achievable goals. Thus, since $p_{\text{skills}}=1$ in ${\cal G}^+$, $p_{\text{goals}}$ is maximal and constant for any area $\omega \in {\cal G}^+ \cap {\cal G}^*$. This implies that the concentration of particles in any area of ${\cal G}^+ \cap {\cal G}^*$ is greater  than if particles were uniformly spread over ${\cal C}$. In that case, for any particle $x$ from the set $\{x_i\}_{i=1}^m$, we know that $P(x \in \omega|KSD(q=\{x_i\}_{i=1}^m,p_{\text{goals}})=0)\geq P(x \in \omega|KSD(q=\{x_i\}_{i=1}^m,U({\cal X}))=0)$, with ${\cal X}$ an hypercube of $d$ dimensions with side length equal to $diam({\cal C})$ and $U({\cal X})$  the uniform distribution over ${\cal X}$. 

Next, if $KSD(q=\{x_i\}_{i=1}^m,U({\cal X}))=0$, we know that particles are spread as a grid over each dimension of ${\cal X}$. Thus, in each dimension of ${\cal X}$ any particle is separated from its closest neighbor by at most a difference of $diam({\cal C})/(\sqrt[d]{m}-1)$ in the worst case. Thus, any area $\omega$ with diameter greater than $\sqrt{d}\frac{diam({\cal C})}{(\sqrt[d]{m}-1)}$  is guaranteed to contain a particle in that case, which concludes the proof. 
\end{proof}
 
 \begin{figure*}[ht]
    \centering
        \includegraphics[width=0.99\textwidth]{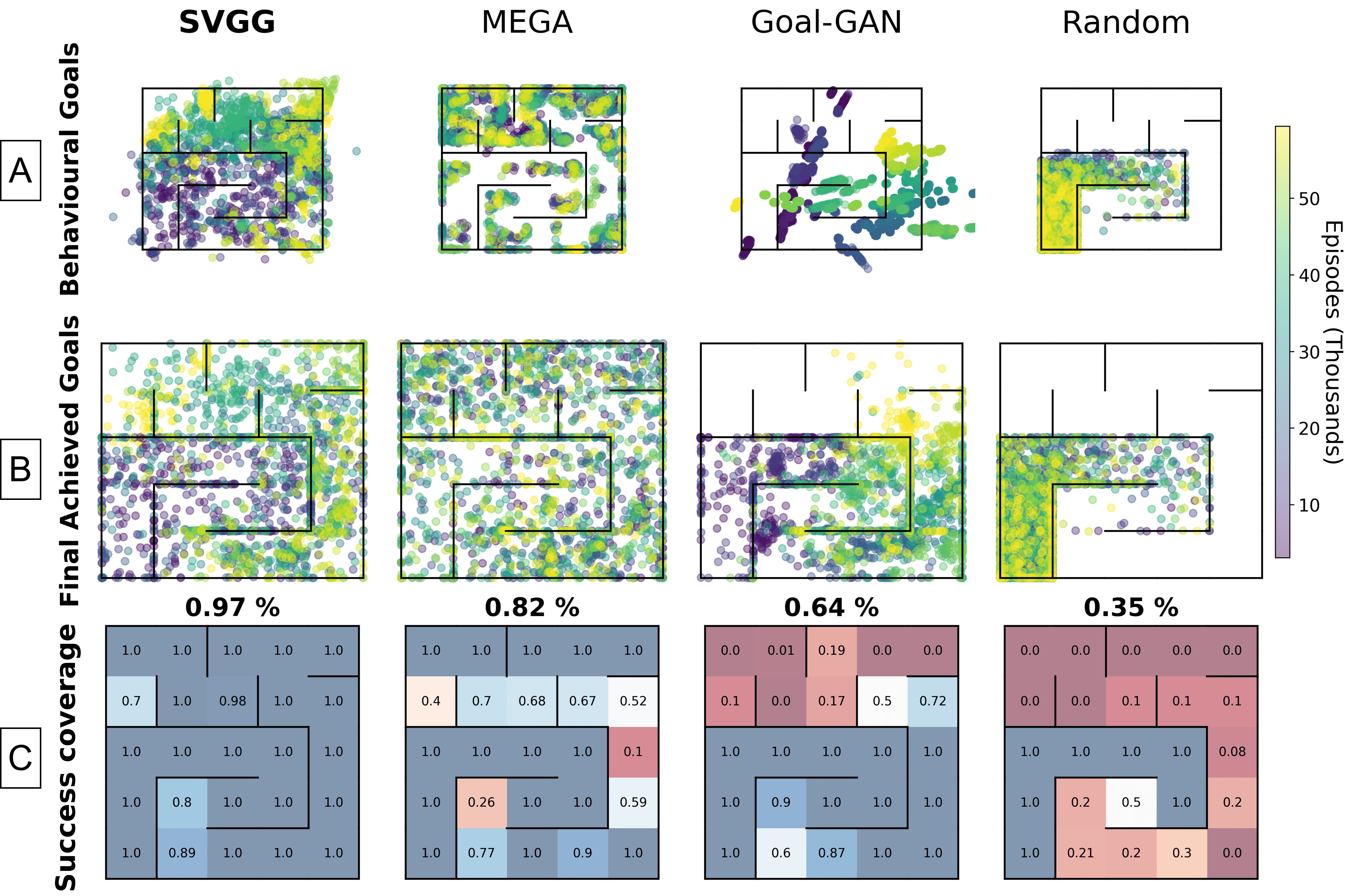}
    \caption{ \label{fig:visu_2}
    Visualization of: (A) behavioral goals , (B) achieved goals and (C) success coverage for 4M steps in Maze 2. \svgg is the only one to achieve nearly perfect success coverage.}
\end{figure*}

\section{Visualization of goals}

We visualize behavioral and achieved goals in Maze 1 (\figurename~\ref{fig:visu_1}) and Maze 2 (\figurename~\ref{fig:visu_2}), in parallel with the success coverage after training. The main advantages of our method lie in the capacity to target difficult areas and avoid catastrophic forgetting, which results in nearly optimal success coverage. We observe that \mega efficiently discovers the environment but fails to master the corresponding goals. This also leads to catastrophic forgetting and a lack of adaptation when the environment changes, as studied in the main paper.

One can see that the generation of GOIDs in \ggan  is very unstable and tricky in such discontinuous goal space, especially as the generator is susceptible to output goals outside the environment boundary, which \svgg avoids with the validity distribution.

\section{Additional experiments details}
\label{app:envs}
 
\subsection{Environments}
 
\paragraph{Pointmaze} We use a 2D pointmaze environment where the state space and goal space are (x, y) coordinates (the agent cannot see the walls), and the agent moves according to its action, which is a 2-dimensional vector with dimensions constrained to $[-0.95, 0.95]$. All methods are tested on four $5\times5$ mazes with different shapes and a highly discontinuous goal space. The maximum number of steps in one trajectory is set to 30. We argue that the difficulty of these environments does not lie in their size, but rather in the complexity of their goal space and thus of the trajectories adopted by the agent to achieve these goals.

\paragraph{Antmaze}
An Ant has to move around in a U-shape hallway maze of size 20x4, the goal space remains $(x,y)$ coordinates of the Ant as in Pointmaze. However, the exploratory behavior is much simpler than in the considered pointmazes environments, as the maze is simply U-shaped. The difficulty lies in the control of the Ant dynamics, as it must move its legs with a 8-dimensional action space and the observation space is a 30-dimensional vector corresponding to the angles of the legs.  

\paragraph{FetchPickAndPlace (Hard version)}

We also perform comparisons on a hard version of FetchPickAndPlace-v1 from OpenAI gym \cite{1606.01540}. The agent controls a robotic arm which must pick and place a block to a 3D desired location. In the hard version, the behavioral goals are all between 20 and 45cm in the air, while in the original version, 50\% of behavioral goals are on the table and the remaining ones are from 0 to 45cm in the air.

While this environment presents a greater difficulty in terms of control (the action dimension is 4 and the state dimension is $24$ which correspond to the various positions of the robot joint), the goal generation component is significantly easier: as the 3-dimensional goal space is smooth, there is no obstacle for the agent to bypass. As \figurename~\ref{fig:main} shows, \svgg solves the environment within 1.5M steps, but does not significantly outperform \mega.

We argue that the interest of our goal generation algorithm resides in environments with highly discontinuous goal space as the action control is largely supported by the RL algorithm (e.g. \ddpg). Therefore, the smaller difference between \mega and \svgg in this environment was expected, as \svgg is mainly designed to target non-smooth goal spaces, and to avoid pitfalls such as catastrophic forgetting.

\paragraph{FetchReach (Stuck version)}
\label{app:fetchreach}

We compare \svgg to baselines with a modified version of the FetchReach environment, where the gripper is initially stuck between four walls and has to carefully navigate between them to reach the goals.
The observations are 9-dimensional vectors, the actions as well as the goals are 3-dimensional.
The success coverage is computed on uniformly sampled goals with a target range fixed to $0.2$, which corresponds to the maximal L2 distance between the goal and the initial gripper position. 
The initial task is solved within $20.000$ steps by all baselines, whereas with the modified version, only \svgg achieves a $0.8$ success rate in 3M steps. 

\paragraph{FetchPush (Maze version)}
\label{app:fetchpush}

We add a maze structure with walls on the table to the FetchPush benchmark to once again add discontinuities in the goal space. The observations are 24-dimensional vectors, the actions as well as the goals are 3-dimensional. The success coverage is computed on goals uniformly sampled on the table, and the block's initial position is sampled behind the first wall on the robot side.
The standard FetchPush task was solved by \svgg and \mega in less than 1M steps. The maze version is much harder: \svgg achieves a $0.8$ success coverage rate within 10M steps, whereas \mega barely reaches $0.5$.
The distance to reach the goals in all environments is $\delta = 0.15$.

\section{Control of the sampling difficulty}
\label{app:alpha_beta}

\begin{figure}[!ht]
    \centering
    \includegraphics[width=\columnwidth]{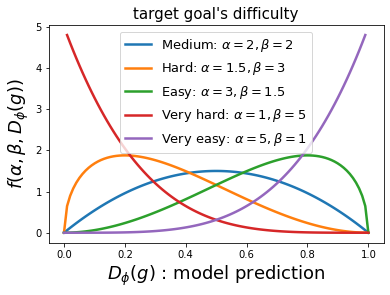}
    \caption{Modulation of goal's target difficulty with beta distributions. $D_{\phi}(g)$ is the model's predicted probability of the agent achieving the goal $g$.}
    \label{fig:betas}
\end{figure}

Using beta distributions makes it possible to tune the goal difficulty an agent should aim at to efficiently explore and control an environment. Indeed, by tuning the $\alpha$ and $\beta$ hyper-parameters of the distribution, one gets a different incentive for goal difficulty. We choose to compare the efficiency of 5 pairs of $\alpha$ and $\beta$ hyper-parameters, as illustrated in \figurename~\ref{fig:betas}. 

\begin{figure*}[!ht]
\centering
    \includegraphics[width=1\textwidth]{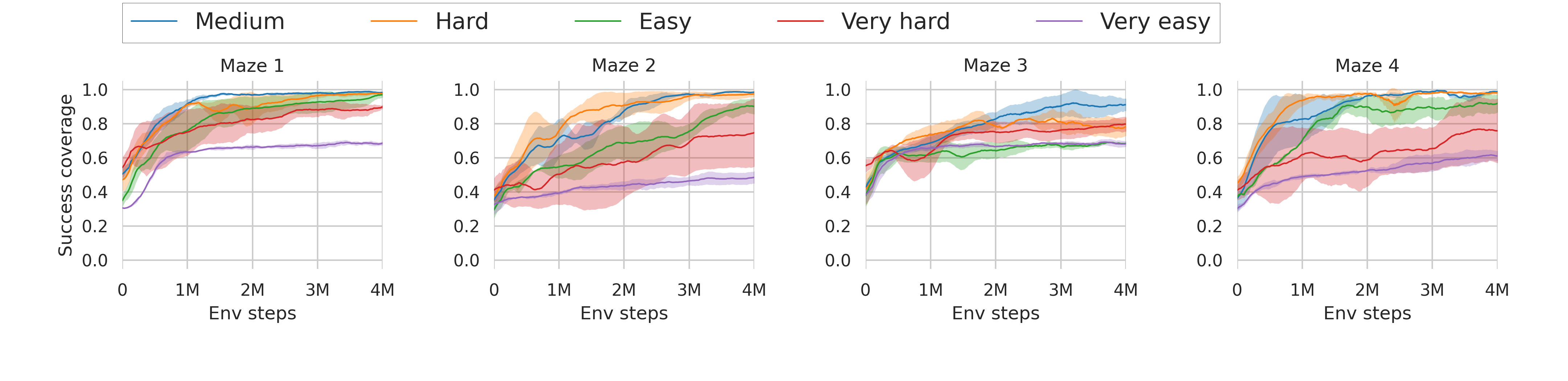}
    \caption{Evolution of success coverage for 5 target difficulties as intrinsic motivation for \svgg, on 4 mazes and 4 seeds each.}
    \label{fig:svgg_study_difficulty_0}
\end{figure*}

In \figurename~\ref{fig:svgg_study_difficulty_0}, we compare the 5 versions of \svgg using different beta distributions from  \figurename~\ref{fig:betas}, on the 4 previously considered mazes. 
One can observe that extreme targets difficulties are the least effective ones, especially the \textit{Very easy}, which is too conservative to efficiently explore new areas of the space.
On the other hand, \svgg performs very well with \textit{Medium} and \textit{Hard} distributions. This suggests that the optimal goal difficulty is somewhere between medium and hard. Performing a more advanced optimization over $\alpha$ and $\beta$ is left for future work.

\section{Target goal distribution ablations}

\begin{figure}[!ht]
\centering
    \includegraphics[width=0.9\columnwidth]{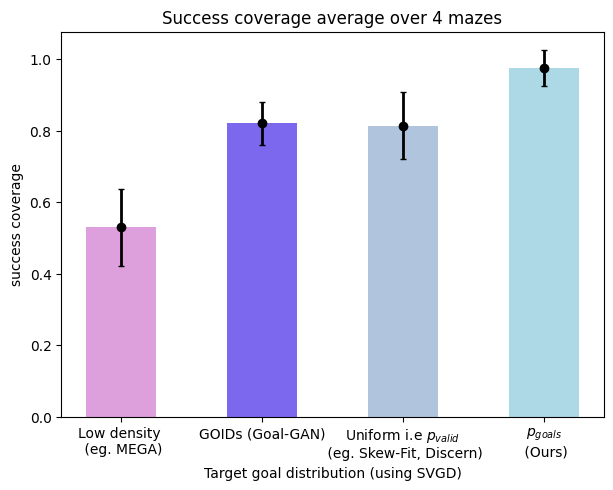}
    \caption{Plot of the average success coverage of \svgg over the 4 mazes (3 seeds each for a total of 12 runs) after 4 millions training steps where we replace our target goal distribution with other choices and keep \svgd as sampling method, the error bar shows the standard deviation between all runs.}
    \label{fig:svgg_target_ablation}
\end{figure}

We conduct an ablation which consists in keeping \svgd, but replacing our target goal distribution (corresponding to goals whose prediction of success is close to 0.5) with the target goal distributions of related work such as of \skewfit, \mega or \ggan (resp. the uniform distribution over the support of achieved goals, the region of low density of achieved goals and goals of intermediate difficulty (GOIDs)). For all criteria, the goal target distribution is represented as a set of particles and optimized with \svgd. We compare the average success coverage over our 4 mazes on \figurename~\ref{fig:svgg_target_ablation} where we can see that our choice of target distribution is the most effective. We describe below the compared goals' distribution.

\paragraph{Uniform distribution over the support of achieved goals (as in \skewfit and \discern):} This corresponds to our "Only validity" baseline in \figurename~\ref{fig:main}. Indeed, the probability for a goal to be valid is learned with a One Class SVM which exactly strives to uniformly cover the distribution of achieved goals. As the results presented in \figurename~\ref{fig:svgg_target_ablation} show, this baseline performs relatively well, but is unable to target areas where the agent struggles. Therefore, it is not as efficient as our target distribution $p_{goals}$.

\paragraph{Distribution derived from the region of low density of achieved goals (as in \mega):} We can obtain a distribution of goals very close to \mega’s by combining our “Only Validity” ablation with a beta distribution tuned to address the lowest probability region (i.e taking $p_{goals} \propto f(\alpha, \beta, V_{\psi}(g))$), with $f$ a Beta distribution and $\alpha$ and $\beta$ set to target low density of $V$). Due to the absence of a mechanism to avoid sampling unfeasible goals, the particles are attracted to low density areas that mostly correspond to non-valid goals, which make this baseline very inefficient.

\paragraph{distribution of GOIDs (as in \ggan):}
Our target distribution $p_{goals}$ is very close to the GOID in \ggan, the main difference is the smoothness of our Beta distribution versus the indicator function of \ggan that labels a goal as of "intermediate difficulty" when the probability of reaching it in an interval (eg. between 0.2 and 0.8). So, to move closer to the \ggan criterion, we replaced the beta-distribution used in \svgg with the crisp distribution (with a generalized Gaussian distribution of skewing parameter $\beta=6$) which outputs 1 for probabilities between 0.3 and 0.7 and 0 otherwise (which are the parameters that give the best results). Note that while this distribution is differentiable, the gradient is less informative than our version. As a consequence, approximating this distribution with \svgd is less efficient and gets a lower success coverage.

\section{Sampling method ablations}

To highlight the interest of the \svgd choice to sample goals from our target distribution $p_{goals}$, we conducted additional experiment where we swap \svgd with some other sampling tools : MCMC (Metropolis Hastings), GANs and direct sampling from the replay buffer.

\begin{figure}[!ht]
\centering
    \includegraphics[width=\columnwidth]{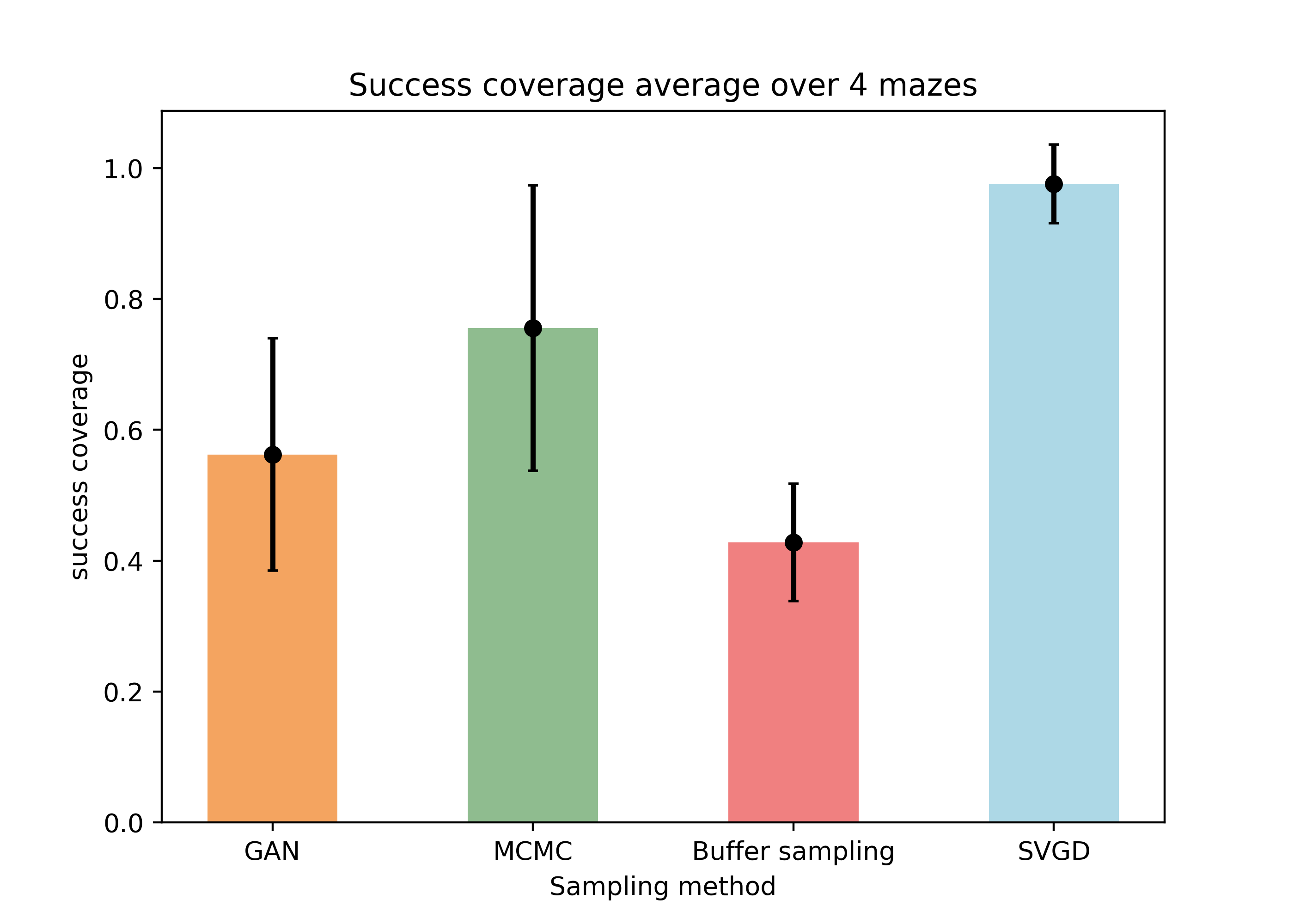}
    \caption{Plot of the average success coverage of \svgg over the 4 mazes (3 seeds each for a total of 12 runs) after 4 millions training steps where we replace \svgd with other sampling methods, the error bar shows the standard deviation between all runs.}
    \label{fig:svgg_sampling_ablation}
\end{figure}

In \figurename~\ref{fig:svgg_sampling_ablation} we can see that \svgd is by far the most efficient way to sample from $p_{goals}$ and thus maximize the success coverage. We describe below the tested sampling tools.

\paragraph{GANs}: We use the same procedure as \ggan by replacing their $GOID$ criterion by our target distribution $p_{goals}$. the results are very similar to \ggan, this can be explained by the proximity of our criterion in terms of intermediate difficulty, except from the fact that we add $p_{valid}$, we can also conclude that GANs are not the best choice for moving target distributions due to their training instability. We use the same hyperparameters as in the \ggan baseline.

\paragraph{Buffer sampling}:
We sample a batch of previously achieved goals in the buffer, and then compute a categorical distribution based on the $p_{goals}$ criterion and sample candidate goals. This method is the least effective baseline, which was expected as the exploration component was provided by \svgd, the goals from the replay buffer are not sufficiently diverse to explore the environment if not combined with a diversity criterion.

\paragraph{MCMC (Metropolis Hastings)}: At each $p_{goals}$ update, we construct a markov-chain with normal movement (with $\mu =0$ and $\sigma=0.5$) then classic likelihood ratio rejection sampling to draw goals samples that are provided to the agent.
Metropolis-Hastings is a good candidate but is still far from the \svgd performance, it presents some good sampling accuracy at times but is very slow to converge to the true distribution, thus the goal sampling isn't always aligned with the agent skills at the time, and enable the recovery property.

%%%%%%%%%%%%%%%%%%%%%%%%%%%%%%%%%%%%%%%%%%%%%%%%%%%%%%%%%%%%%%%%%%%%%
\section{Methods details}
\label{app:methods}

We first focus on the behavioral goal sampling strategy. We use \ddpg with \her to learn how to reach a goal in \svgg, \mega and \ggan.
\ddpg is parameterized as in Table~\ref{param:ddpg}. All algorithms are run using the same number of training epochs $n$.

\begin{table}[H]
\centering
\caption{\ddpg parameters\label{param:ddpg}}
\begin{tabular}[t]{lcc}
\hline
\textbf{DDPG Hyper-Parameters}&\textbf{Value}\\
\hline
\\
Batch size for replay buffer & 2000\\
Discount factor $\gamma$ &0.99\\
Action L2 regularization &0.1\\
(Gaussian) Action noise max std  &0.1\\
Warm up steps before training  &2500\\
Actor learning rate &1e-3\\
Critic learning rate &1e-3\\
Target network soft update rate&0.05\\
Actor \& critic networks activation & Gelu \\
Actor \& critic hidden layers sizes & $512^3$ \\
Replay buffer size & nb training steps \\
\hline
\end{tabular}
\end{table}%

\subsection{Hindsight Experience Replay}
  
The original goal relabeling strategy introduced in \her \cite{andrychowicz2017hindsight} is the \textit{future}, which consists in relabeling a given transition with a goal achieved on the trajectory later on. This is very effective in sparse reward setting to learn a GCP. However, many works suggested that relabeling transitions with goals outside the current trajectory helps the agent generalize across trajectories. For example, one can use inverse RL to determine the optimal goal to relabel a given transition \cite{eysenbach2020rewriting}. We use a naive version of this method. As in \cite{pitis2020maximum}, we relabel transitions for \ddpg optimization using a mixed strategy. All methods presented in this work use the same strategy.
  
10\% of real experience are kept while 40\% of the transitions are relabeled using the \textit{future} strategy of \her. We relabel the remaining 50\% transitions with goals outside of their trajectory, with randomly sampled goals among the past behavioral and achieved goals. The latter part of the strategy helps share information between different trajectories that often contains similar transitions.   

\subsection{SVGG}
\label{sec:svgg_algo}
\svgg is described in the main paper, in this section, we go through implementation details and hyper-parameters.

\begin{algorithm*}[ht]
\caption{\textit{RL with Stein Variational Goal Generation}}
\label{algo:svgg}
\begin{algorithmic}[1]
\STATE \textbf{Input:} a GCP $\pi_\theta$, % with parameters $\theta$,
%a parameterizable environment ${\cal M}$, 
a success predictor $D_\phi$, % with parameters $\phi$,
a reachability predictor $R_\psi$, 
buffers of transitions ${\cal B}$, reached states ${\cal R}$ and success outcomes $O$, a kernel $k$, hyper-parameters $t^{(r)}$, $t^{(m)}$, $t^{(p)}$ standing for number of rollouts, model updates and particles updates respectively, and hyper-parameter $b$  standing for the  batchsize of the model update.   
%\STATE Set $p_c$ as the uniform distribution on $C$;
\STATE Sample a set of particles $q=\{x^i\}_{i=1}^m$ uniformly from states reached by pre-runs of $\pi_\theta$; % in a sequence of pre-runs using random goals; 
\FOR{$n$ epochs}  %\\[0.2cm]
%$\qquad \qquad \qquad \qquad \qquad \qquad \qquad \qquad \qquad \quad$$\vartriangleright$ \textit{Model Update} 
%\STATE blabla \COMMENT{\textit{Data Collection}} 
%\Statex{\hspace{1cm} \textit{\#\#\#\#\#\#\# Data Collection \#\#\#\#\#\#\#}}
\FOR[$\qquad \qquad \qquad \qquad \qquad \qquad \qquad \qquad \quad \qquad$  $\vartriangleright$
\textit{Data Collection}]{$t^{(r)}$ iterations}
\STATE Sample a goal g from $q$; % batch %of $l^{(r)}$ particles
%$\{g^i\}_{i=1}^{l^{(r)}}$ %uniformly
%from $\Omega$; 
%\FOR{$i$ from $1$ to $l^{(r)}$} % \in [\![1;l]\!]$} 
\STATE $\tau \leftarrow$ Rollout $\pi_\theta(.|.,g)$;
\STATE Store all $(s_t,a_t,s_{t+1},r_{t},g)$ from $\tau$ in ${\cal B}$ and every $s_t$ from $\tau$ in ${\cal R}$;
\STATE Optionally (\her): Store relabeled transitions from $\tau$;
\STATE Store outcome $(g,I(s_{|\tau|} \approx g))$ in $O$;  %Store all states from $\tau$ in {\cal R}; 
\label{store1A} 
%\STATE Optionally: Store $(s_{|\tau|},1)$ in ${\cal A}$; 
%\ENDFOR    
\ENDFOR 
%\\[0.2cm] 
%\Statex {\hspace{1cm} \textit{\#\#\#\#\#\#\# Model Update \#\#\#\#\#\#\#}}
\FOR[$\qquad \qquad \qquad \qquad \qquad \qquad \qquad \qquad \qquad \quad$$\vartriangleright$ \textit{Model Update}]{$t^{(m)}$ iterations} 
\STATE Sample a batch %of $l^{(m)}$ outcomes \\ $\qquad 
$\{(g^i,s^i)\}_{i=1}^{b}$ %uniformly
from $O$;
\STATE Oversample the minority class w.r.t. $s$ to get a balanced success/failure dataset;
\STATE Update model $D_\phi$ (e.g., via ADAM), with gradient of ${\cal L}_\phi$ \eqref{model_loss});  
\ENDFOR %\\[0.2cm] 
%\Statex {\hspace{1cm} \textit{\#\#\#\#\#\#\# Prior  Update \#\#\#\#\#\#\#}}
\STATE Update model $R_\psi$ according to all states in ${\cal R}$;  $\qquad \qquad \quad \qquad \qquad$$\vartriangleright$ \textit{Prior  Update}
%\Statex {\hspace{1cm} \textit{\#\#\#\#\#\#\# Particles Update \#\#\#\#\#\#\#}}
%\STATE Compute $K$ as the Gram matrix  of particles in $\Omega$ using kernel $k(.,.)$;
%\STATE Compute tensor $R=[\nabla_{x_i} k(x_i,x_j)]_{(x_i,x_j) \in \Omega^2}$;
%\STATE Compute $D=[\log(\dfrac{D_\phi(x_i)}{1-D_\phi(x_i)}) \nabla_{x_i}D_\phi(x_i)]_{x_i \in \Omega}$;
\FOR[$\qquad \qquad \qquad \qquad \qquad \qquad \qquad \qquad \qquad \quad$ $\ \vartriangleright$ \textit{Particles Update}]{$t^{(p)}$ iterations} 
\STATE Compute the density of the target $p_{\text{goals}}$ for the set of particles $q$ using \ref{SVGG_distrib}; 
%\STATE $\{{p_{\text{goals}}(x_i)}\}_{i=1}^n = \{{p_{\text{skills}}(x_i) . p_{\text{prior}}(x_i)}\}_{i=1}^n$
\STATE Compute transforms: % for every particle $i$ from $\Omega$:  %$\{\phi^*(x_i)\}_{i=1}^n$
%\STATE
$\phi^*(x_i) =\frac{1}{m}\sum_{j=1}^m \big [  k(x_j,x_i) \nabla_{x_j} \log p_{\text{goals}}(x_j) + \nabla_{x_j} k(x_j,x_i) \big ]$; %equivalent to \eqref{optimal_transform}
\STATE Update particles $x_{i} \leftarrow x_{i} + \epsilon \phi^*(x_i), \quad \forall i=1 \cdots m$;
%\STATE Sample a batch $\{x^i\}_{i=1}^{l^{(p)}}$ from $\Omega$;
%\STATE Update particles from the batch using $K$, $R$ and $D$ , equivalent to \eqref{particle_moves2} ; 
\ENDFOR
%\\[0.2cm]
\STATE Improve agent with any Off-Policy RL algorithm  $ \qquad \qquad \qquad \qquad $%\hspace{1cm} 
$\vartriangleright$
 \textit{Agent Improvement}
\STATE (e.g., \ddpg) using transitions from $\cal B$;
 %\Statex{\hspace{1cm} (e.g., \ddpg) using transitions in $\cal B$;}
\ENDFOR
%\Return{ } %$\pi_\theta$
\end{algorithmic}
\end{algorithm*}

\begin{table}[H]
\centering
\begin{tabular}[t]{lccc}
\hline
\textbf{SVGG Hyper-Parameters}&\textbf{Symbol}&\textbf{Value}\\
\hline
\\
\textbf{SVGD} &  \\
Number of particles& $m$ & 100 \\
Optimization interval (in steps)& & 20 \\
Nb of particle moves per optim. &$k^{(p)}$ & 1 \\
RBF kernel $k(.,.)$ bandwidth & $\sigma$ & 1 \\
Learning rate & $\epsilon$ & 1e-3 \\
\hline
\\
\textbf{Agent's skill model $D_{\phi}$} &  \\
Hidden layers sizes & & (64, 64) \\
Gradient steps per optimization & $K$ & 10 \\
Learning rate & & 1e-3 \\
Training batch size & $l^{(m)}$ &100\\
Training history length (episodes) & &1000\\
Optimization interval (in steps) & & 4000 \\
Nb of training steps & $k^{(m)}$ & 100 \\
Activations & & Gelu \\
\hline
\\
\textbf{OCSVM validity distribution} &  \\
RBF kernel $k(.,.)$ bandwidth &$\sigma$  & 1 \\
Optimization interval (in steps) & & 4000 \\
\hline
\end{tabular}
\end{table}%

\subsection{MEGA}
 
The authors of \mega \cite{pitis2020maximum} train a GCP with previously achieved goals from a replay buffer. Their choice of goals relies on a minimum density heuristic, where they model the distribution of achieved goals with a KDE. They argue that aiming at novel goals suffices to efficiently discover and control the environment. We use the original implementation of the authors, the pseudocode is given in Algorithm~\ref{algo:MEGA} and specific hyper-parameters are in Table~\ref{params:mega}.

\begin{algorithm*}[!ht]
\caption{\textit{MEGA}}
\label{algo:MEGA}
\begin{algorithmic}[1]
\STATE \textbf{Input:} a GCP $\pi_\theta$, 
%a parameterizable environment ${\cal M}$, 
buffer of reached states ${\cal R}$, 
a KDE model $P_{as}$ of the achieved states,  
numbers $c$, $N$, $t^{(r)}$ and $l^{(r)}$.
\STATE Initialize ${\cal R}$ with states reached by pre-runs of $\pi_\theta$; % in a sequence of pre-runs using random goals; 
\FOR[$\qquad \qquad \qquad \qquad \qquad \qquad \qquad \qquad \qquad \qquad \qquad$  $\vartriangleright$
\textit{Data Collection} ]{$n$ epochs}%\\[0.2cm]

\FOR{$t^{(r)}$ iterations}
\STATE Sample a batch %of $l^{(r)}$ particles
$\{g^i\}_{i=1}^{N}$ uniformly from $\cal{R}$;
\STATE Eliminate unachievable candidates if their $Q$-values are below cutoff $c$;
\STATE Choose goals $\{g^i\}_{i=1}^{l^{(r)}} = \argmin_{g_i} P_{as}(g_i)$
\FOR{$i$ from $1$ to $l^{(r)}$} 
\STATE $\tau \leftarrow$ Rollout $\pi_\theta(.|.,g=g^i)$;
\STATE Store all $(s_t,a_t,s_{t+1},r_{t},g^i)$ from $\tau$ in ${\cal B}$ and every $s_t$ from $\tau$ in ${\cal R}$;
\STATE Optionally (\her): Store relabeled transitions from $\tau$; 
\STATE Store outcome $(g^i,I(s_{|\tau|} \approx g^i))$ in $O$;  %Store all states from $\tau$ in {\cal R}; 
%\label{store1A} 
%\STATE Optionally: Store $(s_{|\tau|},1)$ in ${\cal A}$; 
\ENDFOR    
\ENDFOR 
%\\[0.2cm] 
%\Statex {\hspace{1cm} \textit{\#\#\#\#\#\#\# Model Update \#\#\#\#\#\#\#}}

\STATE Update KDE model $P_{as}$ with uniform sampling from ${\cal R}$;  $ \qquad \qquad \qquad \  $$\vartriangleright$ \textit{Model Update} 

\STATE Improve agent with any Off-Policy RL algorithm  {$ \qquad \qquad \qquad \qquad $%\hspace{1cm} 
$\vartriangleright$
 \textit{Agent Improvement}}
\STATE (e.g., \ddpg) using transitions from $\cal B$;
\ENDFOR
%\Return{ } %$\pi_\theta$
\end{algorithmic}
\end{algorithm*}

\begin{table}[H]
\centering
\caption{\mega parameters\label{params:mega}}
\begin{tabular}[t]{lccc}
\hline
\textbf{MEGA Hyper-parameters}&\textbf{Symbol}&\textbf{Value}\\
\hline
\\
RBF kernel bandwidth & $\sigma$ & 0.1 \\
KDE optimization interval (in steps) && 1 \\
Nb of state samples for KDE optim. && 10.000 \\
Nb of sampled candidate goals &$N$ & 100 \\
$Q$-value cutoff & $c$ & -3\\
\hline
\end{tabular}
\end{table}%

\subsection{GoalGAN}
 
\begin{algorithm*}[!ht]
\caption{\textit{GoalGAN}}
\label{algo:goal-gan}
\begin{algorithmic}[1]
\STATE \textbf{Input:} a GCP $\pi_\theta$, 
a goal Generator $G_{\theta_g}$, a Discriminator $D_{\theta_d}$,
a success predictor $D_\phi$,
%a convex subset $C \subseteq S$ containing all reachable states in $\cal M$,
buffers of transitions ${\cal B}$, %a buffer of
reached states  ${\cal R}$ and  %a buffer of
success outcomes $O$, numbers $t^{(r)}$, $l^{(r)}$, $l^{(m)}$, $k^{(m)}$, $l^{(g)}$ and $k^{(g)}$.  
%\STATE Set $p_c$ as the uniform distribution on $C$;
\STATE Initialize  $G_{\theta_g}$ and $D_{\theta_d}$ with pre-runs of $\pi_\theta$; % in a sequence of pre-runs using random goals; 
\FOR[$\qquad \qquad \qquad \qquad \qquad \qquad \qquad \qquad \qquad \qquad \qquad$  $\vartriangleright$
\textit{Data Collection}]{$n$ epochs}%\\[0.2cm]
%\Statex{\hspace{1cm} \textit{\#\#\#\#\#\#\# Data Collection \#\#\#\#\#\#\#}}
\FOR{$t^{(r)}$ iterations}
\STATE Sample noise $\{z_i\}_{i=1}^{l^{(r)}} \sim \mathcal{N} (0,1)$
\STATE generate $\{g^i\}_{i=1}^{l^{(r)}} = G_{\theta_g}(\{z_i\}_{i=1}^{l^{(r)}})$
\FOR{$i$ from $1$ to $l^{(r)}$} % \in [\![1;l]\!]$} 
\STATE $\tau \leftarrow$ Rollout $\pi_\theta(.|.,g=g^i)$;
\STATE Store all $(s_t,a_t,s_{t+1},r_{t},g^i)$ from $\tau$ in ${\cal B}$ and every $s_t$ from $\tau$ in ${\cal R}$;
\STATE Optionally (\her): Store relabeled transitions from $\tau$; 
\STATE Store outcome $(g^i,I(s_{|\tau|} \approx g^i))$ in $O$;  %Store all states from $\tau$ in {\cal R}; 
%\label{store1A} 
%\STATE Optionally: Store $(s_{|\tau|},1)$ in ${\cal A}$; 
\ENDFOR    
\ENDFOR 
%\\[0.2cm] 
%\Statex {\hspace{1cm} \textit{\#\#\#\#\#\#\# Model Update \#\#\#\#\#\#\#}}
 
\STATE Sample a batch $\{(g^i,s^i)\}_{i=1}^{l^{(g)}}$ 
from $O$; $\qquad \qquad \qquad \qquad \qquad \qquad$ $\vartriangleright$ \textit{GAN training}
\STATE Label goals (GOID or not) with model $D_\phi$ : $\{y_{g_i}\}_{i=1}^{l^{(g)}} = \{P_{min} < D_\phi(g_i) < P_{max}\}_{i=1}^{l^{(g)}}$
\FOR{$k^{(g)}$ iterations}
\STATE Update $G_{\theta_g}$ and $D_{\theta_d}$ (e.g. with ADAM) with gradients of LSGAN losses; \eqref{GAN_losses}
\ENDFOR %\\[0.2cm] 

\FOR[$\qquad \qquad \qquad \qquad \qquad \qquad \qquad \qquad \qquad \quad$ $\vartriangleright$ \textit{Model Update} ]{$k^{(m)}$ iterations} 
\STATE Sample a batch %of $l^{(m)}$ outcomes \\ $\qquad 
$\{(g^i,s^i)\}_{i=1}^{l^{(m)}}$ %uniformly
from $O$; 
\STATE Update model $D_\phi$ (e.g., via ADAM), with gradient of ${\cal L}_\phi$ (agent model loss described in the paper); %gradient descent steps: %on a binary cross entropy loss:
%$\nabla_\phi {\cal L}_\phi =  \sum_{(g^i,s^i) \in O} s^i \nabla_\phi (\log D(g^i)) + (1-s^i) \nabla_\phi (\log (1-D(g^i)))$ 
\ENDFOR %\\[0.2cm] 
\STATE Improve agent with any Off-Policy RL algorithm  {$ \qquad \qquad \qquad \qquad $%\hspace{1cm} 
$\vartriangleright$
 \textit{Agent Improvement} }
\STATE (e.g., \ddpg) using transitions from $\cal B$; 
\ENDFOR
\end{algorithmic}
\end{algorithm*}

\ggan \cite{florensa2018automatic} uses a procedural goal generation method based on GAN training. As our \svgg, it aims at sampling goals of intermediate difficulty, which they define as $\mathcal{G}_\text{GOID} = \{g | P_{min}< P_{\pi}(g) < P_{max}\}$, $P_{\pi}(g)$ being the probability for the policy $\pi$ to achieve goal $g$, $P_{min}$ and $P_{max}$ are hyper-parameters. A Discriminator $D_{\theta_d}$ is trained to distinguish between goals in $\mathcal{G}_\text{GOID}$ and other goals, while a generator $G_{\theta_g}$ is trained to output goals in $\mathcal{G}_\text{GOID}$ by relying on the discriminator outputs. They optimize $G_{\theta_g}$ and $D_{\theta_d}$ in a manner similar to the Least-Squares GAN (LSGAN) with the following losses:
 
\begin{equation}
\label{GAN_losses}
\begin{split}
    \min_{\theta_d} V(D_{\theta_d}) =& \mathbb{E}_{g \sim \cal{R}}\big [y_g (D_{\theta_d}(g) - 1)^2 \\
    &+(1-y_g)(D_{\theta_d}(g)+1)^2 \big ] \\
    \min_{\theta_g} V(G_{\theta_g}) =& \mathbb{E}_{z \sim \mathcal{N}(0,1)} \big[ D_{\theta_d}(G_{\theta_g}(z))^2 \big],
\end{split}
\end{equation}

where $y_g$ is the label that indicates whether $g$ belongs to $\mathcal{G}_\text{GOID}$ or not. In \cite{florensa2018automatic}, the authors  use Monte-Carlo sampling of the policy to estimate $y_g$. For efficiency reasons, we use a learned model of the agent's capabilities as in \svgg. The pseudocode is given in Algorithm~\ref{algo:goal-gan} and specific hyper-parameters are in Table~\ref{params:ggan}.

\begin{table}[!ht]
\centering
\caption{\ggan parameters\label{params:ggan}}
\begin{tabular}[t]{lccc}
\hline
\textbf{GoalGAN Hyper-parameters}&\textbf{Symbol}&\textbf{Value}\\
\hline
\\
Gaussian prior dimension && 4 \\ 
Generator hidden layers sizes && (64, 64) \\
Discriminator hidden layers sizes && (64, 64) \\
Optimization interval (in steps) && 2000 \\
GAN training batch size &$l^{(g)}$ & 200 \\
Nb of GAN optimization steps &$k^{(g)}$ & 100\\
GAN Learning rate && 1e-3 \\
Minimum GOID probability &$P_{min}$& 0.1 \\
Maximum GOID probability &$P_{max}$& 0.9 \\
\hline
\end{tabular}
\end{table}%

\subsection{Ressources} Every seed of each experiment was run on 1 GPU in the following list \{Nvidia RTX A6000, Nvidia RTX A5000, Nvidia TITAN RTX, Nvidia Titan Xp, Nvidia TITAN V\}. The total training time is close to 4.000 hours, most of it executed in parallel, as we had full time access to 12 GPUs. 

\end{document}